\documentclass{article} 
\usepackage{main,times}

\usepackage{url}

\usepackage{enumitem}
\usepackage{graphicx}
\usepackage{amsmath,amssymb,amsthm}
\usepackage{multirow}
\usepackage{booktabs}
\usepackage{threeparttable}

\title{Scalable Hierarchical Graph Generation via Soft Community Structure}

\author{Ahmet T{\"u}zen, Helge Langseth \& Kjetil N{\o}rv{\aa}g \\
Department of Computer Science \\ Norwegian University of Science and Technology\\
\texttt{\{ahmet.tuzen,helge.langseth,noervaag\}@ntnu.no}
}

\newcommand{\bA}{\mathbf{A}}
\newcommand{\bx}{\mathbf{x}}
\newcommand{\bX}{\mathbf{X}}
\newcommand{\Apool}{\bA_{\mathrm{p}}}
\newcommand{\xpool}{\bx_{\mathrm{p}}}
\newcommand{\bS}{\mathbf{S}}
\newcommand{\Gen}{\hat{G}}
\newcommand{\bz}{\mathbf{z}}
\newcommand{\bW}{\mathbf{W}}

\newcommand{\method}{\textsc{Schema}}

\iclrfinalcopy
\begin{document}

\maketitle

\begin{abstract}
Generating large attributed graphs requires reproducing the topology, generating attributes jointly with the structure, and remaining scalable. Many real-world graphs exist as a single large graph, so a generative model has to generalize from the one graph it is fit on, without independent samples. We present \method{}, which recursively decomposes a reference graph into a hierarchy of soft communities, assigning each node a membership distribution. Generation is then split into three stages, each trained independently: (1) synthesizing node attributes conditioned on soft memberships, (2) generating intra-community edges from local structural context, and (3) modeling inter-community connections over bridge nodes whose membership mass is distributed across several communities. No stage forms the full adjacency matrix, and each stage operates on a subgraph bounded by the community size. We also introduce an evaluation protocol that covers structural fidelity, memorization, downstream utility, and scalability. On four real-world attributed graphs, \method{} recovers the balance between local and long-range structure more closely than any other model that generates attributes, while reproducing only a small fraction of the reference edges. It retains the downstream accuracy of the reference graph without raising it artificially above that level. Baselines that match its structural fidelity memorize the reference, while those with higher downstream accuracy either exceed the reference accuracy or fail to complete on the larger graphs. We measure scalability on six additional graphs with up to 10 million nodes.
\end{abstract}


\section{Introduction}
\label{sec:intro}

Graphs represent entities and their relationships across many real-world domains, including molecular discovery, social network analysis, and financial fraud detection. Access to real graph data is often restricted by privacy regulations, licensing, or data collection costs, which creates a need for synthetic graphs. 

Graph generation methods fall into two categories, distinguished by the available data and the generation objective. The first category targets collections of relatively small graphs, where a model learns a distribution over independent graph instances and samples new graphs from it, as in molecular generation. The second category targets a single large graph, such as a social, citation, or transaction network, which may contain millions of nodes. In this setting, the model does not sample from a distribution over graph instances; it generates a graph that preserves the structural and attribute characteristics of the reference while generalizing beyond the specific edges and node identities observed during training.

Because the two categories rely on fundamentally different model assumptions, direct transfer between them is impractical. Methods designed for collections of small graphs assume graph-level independence, bounded graph sizes, and access to many training instances, and none of these assumptions hold when a single graph is the only available reference. Learning generative patterns from a single observed graph creates two failure modes that a model must avoid simultaneously: overfitting to the observed edges and node identities, and failing to capture structural characteristics that only become apparent at larger scales, such as community organization and long-range connectivity. Methods developed specifically for this setting avoid the assumptions of the first category but face three limitations of their own.

The first limitation is scalability: autoregressive methods decode edges sequentially, incurring a cost that typically scales quadratically with the number of nodes~\citep{GraphRNN}, whereas methods that operate on full adjacency representations or require repeated passes over the entire graph become memory-limited as the graph grows. The second limitation concerns the coexistence of multiple structural scales. Real-world graphs typically contain dense local communities, sparse inter-community connections, and global topological dependencies~\citep{fortunato}, and generative fidelity depends on preserving all three at once. A single flat decoder has to fit these different densities with one set of parameters, so fitting one density can come at the cost of the others. The third limitation concerns node features. Several existing methods generate only the topology and omit node features~\citep{NetGAN, CELL}, which restricts their use in downstream tasks that operate on attributed graphs, such as node classification, anomaly detection, and recommendation.

Evaluation in this setting is itself underdeveloped. Distributional metrics such as Maximum Mean Discrepancy (MMD)~\citep{MMD} require samples from a reference distribution and do not generalize to a setting with a single reference graph. Existing benchmarks, therefore, often report narrow fidelity metrics and leave open the question of whether a generated graph merely memorizes the reference or remains useful for downstream tasks.

Large real-world graphs are organized into communities, and a generation method can exploit this organization to reduce the size of each subproblem it solves. Following this observation, we present \method{} (\textbf{S}oft \textbf{C}ommunity \textbf{H}ierarchies for \textbf{E}fficient \textbf{M}assive \textbf{A}ttributed graph generation), which recursively decomposes the reference graph into a hierarchy of communities. The decomposition assigns soft labels: each node carries a distribution over community memberships rather than a single label. Nodes whose membership mass is distributed over several communities are called \textit{bridge nodes}, and the generated graph routes its inter-community edges through them. Generation is factorized into three stages, each operating at the community level. The intra-community and inter-community edge generators specialize in different structural densities: dense structure within a community and sparse structure between communities, respectively. Node features and edges are conditioned on the same soft memberships, so the attributes and the connectivity are tied to the same decomposition.

The main contributions of this work are summarized as follows:

\begin{itemize}[nosep]
    \item \textbf{Hierarchical factorization.} We formulate attributed graph generation as a hierarchical factorization over recursively constructed soft communities, decomposed into node feature generation, intra-community edge generation, and inter-community edge generation. Each stage operates on a subgraph bounded by the community size and does not form the full adjacency matrix.
    
    \item \textbf{Soft communities and bridge nodes.} We introduce a learning-free soft-community representation obtained by propagating community memberships over the graph, and an inter-community edge generator that weights candidate pairs by the connectivity observed between communities, so that bridge nodes carry the inter-community structure that hard partitioning removes.
    
    \item \textbf{Evaluation protocol.} We introduce an evaluation protocol that covers structural fidelity, memorization and generalization, downstream utility, and scalability.
    
    \item \textbf{Empirical study.} We evaluate \method{} against nine baselines on attributed graphs ranging from a few thousand to 10 million nodes. Among the models that generate attributes, it is the only one that is competitive on structure, keeps memorization low, preserves downstream accuracy without artificially raising it above the reference graph, and completes on every graph.  
\end{itemize}


\section{Related Work}
\label{sec:related}

We review related work in four areas: graph generation methods developed for the multi-graph setting, methods designed for the single-graph setting, hierarchical and subgraph-based approaches, and evaluation practice.

\subsection{Generation over Graph Collections} 

Most existing graph generation methods are designed for the multi-graph setting, where models are trained on collections of relatively small graphs and generate new samples from the learned distribution. Early methods are largely autoregressive, constructing graphs by sequentially predicting nodes and edges conditioned on the previously generated components. GraphRNN~\citep{GraphRNN} models the process at two levels to capture both global structure and local edge dependencies, BiGG~\citep{BiGG} improves scalability for large sparse graphs through a binary tree representation of the adjacency, and GRAN~\citep{GRAN} generates blocks of nodes and their associated edges at each step instead of one node at a time.

Non-sequential approaches include GraphVAE~\citep{GraphVAE}, which reconstructs a probabilistic graph representation in a single step with a variational autoencoder, and MolGAN~\citep{MolGAN}, which combines a generative adversarial objective with reinforcement learning over chemical properties. Diffusion models have substantially improved generation quality. DiGress~\citep{DiGress} introduces a discrete denoising diffusion model over graphs with categorical node and edge attributes, GraphGDP~\citep{GraphGDP} generates graphs through a continuous-time diffusion process guided by position-enhanced score functions, and EDGE~\citep{EDGE} restricts diffusion to a sparse set of active edges and conditions on the degree sequence to reach larger graphs.

These methods differ substantially in their generative mechanisms but share the assumption that a collection of training graphs is available, and their objective is to learn a distribution over graphs rather than to reproduce the properties of a single observed one. Even the methods that address scalability within this setting are typically demonstrated on graphs with only a few thousand nodes.

\subsection{Generation from a Single Graph} 

Classical models generate graphs from a small number of parameters fit to the reference, such as edge-independent~\citep{ER}, small-world~\citep{WS}, and preferential attachment~\citep{BA} models. The configuration model~\citep{configuration} reproduces the degree sequence exactly by construction, at the cost of all higher-order structure, while DC-SBM~\citep{DC-SBM} fits community-level connectivity but remains constrained by its block-level parametric form. GAE and VGAE~\citep{VGAE} were proposed for link prediction rather than generation, but are commonly used as generative baselines by sampling edges from the decoder applied to the learned embeddings. NetGAN~\citep{NetGAN} models graph generation as learning a distribution over biased random walks with a generative adversarial network, and reconstructs the edge set from sampled walks. Building upon NetGAN, CELL~\citep{CELL} reformulates this random walk objective as a low-rank matrix approximation problem, improving training speed. None of these methods generates node features.

Among the methods that generate attributes alongside structure, GraphMaker~\citep{GraphMaker} employs discrete diffusion in two variants that differ in whether attributes and edges are denoised jointly or asynchronously, with edge mini-batching for scalability. CGT~\citep{CGT} models per-node computation graphs with a Transformer and does not generate a graph directly. GenCAT~\citep{GenCAT} takes an algorithmic approach, reproducing the joint distribution of node classes, attributes, and topology via user-specified characteristic matrices that explicitly control the class-attribute and class-topology relationships. SynGen~\citep{SynGen} generates structure with a generalized stochastic Kronecker model and features with a separate model, and then maps features onto the generated structure with a gradient-boosting aligner intended to preserve feature-structure correlation.

\subsection{Hierarchical and Subgraph-Based Generation} 

Hierarchical and subgraph-based approaches decompose graphs into smaller units to control the size of each generation subproblem, either by constructing an explicit hierarchy or by sampling subgraphs from the reference graph. In the multi-graph domain, HiGen~\citep{HiGen} constructs coarsening hierarchies, while HiCGG~\citep{HiCGG} trains community-level generators integrated via graph encoders. In the single-graph setting, SaGess~\citep{SaGess} partitions the graph into manageable subgraphs and generates their edges with a sampling procedure built on DiGress~\citep{DiGress}. LGSG~\citep{LGSG} extends SaGess with learned embeddings in place of node identifiers, and aggregates the generated subgraphs through embedding distance, not shared identifiers. SANGEA~\citep{SANGEA} partitions an attributed graph into subcommunities with the Louvain algorithm, trains independent generative models for each partition, and links them to preserve topology and features. 

\method{} differs from these methods in three ways. It decomposes the graph recursively, not into a flat set of partitions. It assigns each node a distribution over communities instead of a single label. It generates connectivity between communities at a dedicated stage, whereas the other methods link independently generated partitions.

\subsection{Evaluation of Graph Generation} 
Evaluation differs between the two settings. In the multi-graph setting, structural fidelity is traditionally assessed using MMD~\citep{MMD} over graph statistics~\citep{GraphRNN}, whose sensitivity to kernel choice is well documented~\citep{obray2022, thompson2022}. In the single-graph setting, evaluation relies on direct statistical comparisons, class homophily metrics~\citep{lim2021, platonov2023}, and downstream utility via synthetic-to-real transferability~\citep{GraphMaker}. A separate line of work establishes memorization as a third dimension of evaluation. \citet{chanpuriya2021} show that edge-independent models cannot reproduce the triangle density of real graphs without high overlap with the reference, and \citet{chanpuriya2024} extend this to a hierarchy of edge, node, and fully dependent models.


\section{Methodology}
\label{sec:method}

\subsection{Overview}

\begin{figure}[t]
\centering
\includegraphics[width=\textwidth]{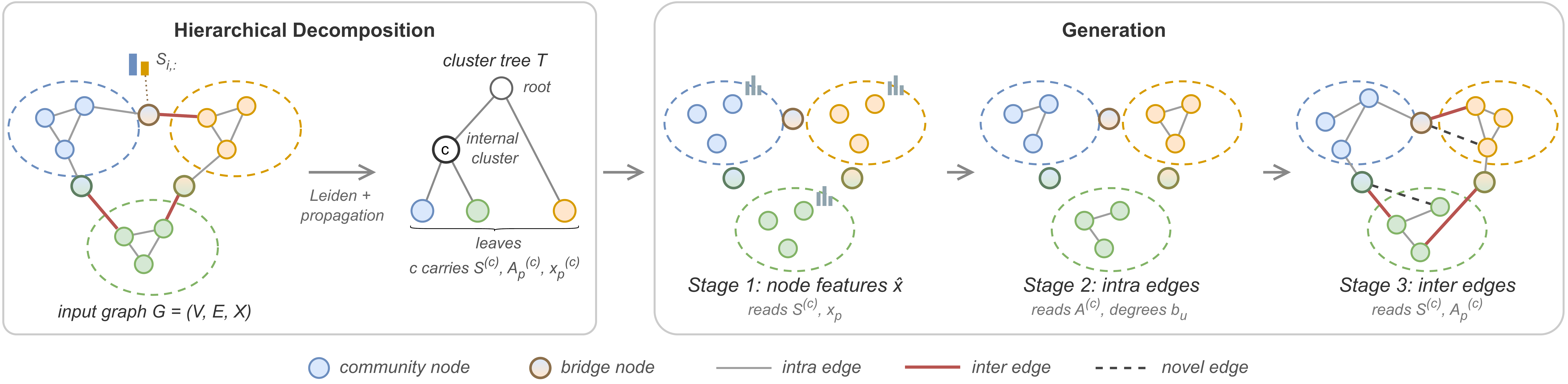}
\caption{Overview of \method. The reference graph is recursively decomposed into soft communities. Generation then proceeds in three stages: node features, intra-community edges, and inter-community edges through the bridge nodes.}
\label{fig:pipeline}
\end{figure}

\method{} consists of a hierarchical decomposition module and a three-stage generator.  The decomposition module takes $G$ and returns a cluster tree $T$ in which every node of $G$ is assigned to a cluster at every level, and in which each internal cluster $c$ carries a soft community assignment $\bS^{(c)}$, a pooled connectivity $\Apool^{(c)}$, and pooled features $\xpool^{(c)}$ (Section \ref{sec:clustering}). The generator reads these artifacts and produces, in order, node features (Stage 1, Section \ref{sec:method-nodes}), edges inside each community (Stage 2, Section \ref{sec:method-intra}), and edges between communities (Stage 3, Section \ref{sec:inter}). The three stages are trained independently and combined in a single traversal of the tree at generation time. The size of each problem is bounded by the community size by design. Figure~\ref{fig:pipeline} illustrates the overall architecture.

\subsection{Preliminaries and Problem Statement}
\label{sec:problem}

An attributed graph is $G = (V, E, \bX)$ with a node set $V$ of $N$ nodes, an edge set $E$, and a feature matrix $\bX \in \mathbb{R}^{N \times F}$ whose $i$-th row holds the features of node $i$. The objective is to generate a new graph $\Gen = (V, \hat{E}, \hat{\bX})$ that reflects the characteristics of $G$. Appendix~\ref{app:anysize} extends this to generated graphs of any size.

An effective generative model must satisfy three requirements. \textbf{Structural fidelity} requires that the structural properties of $\Gen$, such as its degree distribution and clustering patterns, match those of $G$. \textbf{Joint attribute generation} requires that attributes and edges are conditioned on the same representation and not sampled independently. \textbf{Generalization} requires that the model generates new graphs and does not reproduce $G$. For example, a naive baseline that inserts or removes a few random edges in $G$ satisfies the first requirement while failing the third.

\subsubsection{Scope and assumptions} 

Three properties of the formulation determine what \method{} generates and what it omits. First, the generated graph is defined over the node set of the reference graph, preserving node identities. This has two consequences. Node labels are inherited through those identities, so \method{} does not model the label distribution, and the shared node set allows memorization measures such as edge overlap. Second, inter-community edges are generated only between bridge nodes, assuming these edges attach to nodes near community boundaries, thereby keeping the candidate set small and allowing this stage to scale. Third, both edge stages score each node pair independently, given only the two nodes and their cluster context, and do not condition on edges that have already been sampled.

\subsection{Hierarchical Soft-Community Decomposition}
\label{sec:clustering}

\subsubsection{Leaf partition and cluster tree}

We partition $G$ once into leaf communities with the Leiden algorithm~\citep{Leiden} under the modularity objective, which assigns a community label $\ell\in\{0,\dots, K_0-1\}$ to every node. These communities are the finest level of the hierarchy and are the leaves of the cluster tree. Each of these communities is then contracted into a supernode, where the weight between two supernodes is the number of reference edges between the corresponding communities,
\begin{equation*}
\label{eq:supernode}
\bA^{\mathrm{super}}_{c,c'} = \sum_{i:\,\ell_i=c}\;\sum_{j:\,\ell_j=c'} \bA_{ij}.
\end{equation*}

\begin{figure}[t]
\centering
\includegraphics[width=\columnwidth]{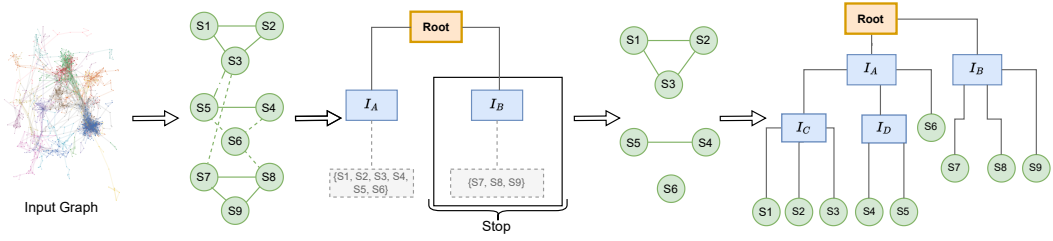}
\caption{Cluster tree construction. Leaf communities are contracted into supernodes, and Leiden is applied recursively to the supernode graph until a termination criterion is met. Here, the branching cap is $b = 3$.}
\label{fig:cluster-tree}
\end{figure}

The supernodes are organized into a tree by applying Leiden recursively to $\bA^{\mathrm{super}}$, processing one set of supernodes at a time and starting from the complete set. A group that still holds several supernodes is split again by the same procedure, while a group holding a single supernode becomes a leaf directly. The recursion terminates on a set that holds at most $b$ supernodes, and every supernode of that set becomes a leaf child of the active tree node. Two cases are handled separately. If the supernodes in the set share no edges, or if Leiden returns the set unpartitioned, the set is split into at most $b$ size-balanced groups by node count, and the recursion continues on the groups that hold more than one supernode. Every leaf of $T$ is one of the communities found in the first partition. Because the splits follow modularity rather than a fixed branching factor, the tree can be unbalanced. Figure \ref{fig:cluster-tree} illustrates the construction.

\subsubsection{Soft assignments via propagation} 

Nodes differ in how strongly they belong to their community. Some lie in its core, and others at the border between two or more communities. Assigning a single label cannot express this difference. We therefore extend the hard labels to soft assignments $\bS^{(c)}$, one matrix per internal cluster $c$, where each row is a distribution over the $K_c$ child clusters of $c$. The row of a node at the core is almost one-hot, and the row of a node at a border carries mass from more than one child cluster. Let $\bS_0$ be the one-hot indicator of the child cluster whose subtree contains each node, and let $\bA^{(c)}$ be the adjacency matrix of the subgraph induced by the nodes under $c$. At the root, this is the adjacency matrix of $G$ itself. The memberships are updated by iterative propagation with row normalization at each step,
\begin{equation}
    \label{eq:propagation}
    \bS_{t+1} = \mathrm{rownorm}\big(\alpha\,\tilde{\bA}\,\bS_t + (1-\alpha)\,\bS_0\big),
\end{equation}
for $t = 0,\dots,\tau-1$, where $\tilde{\bA}$ is the symmetrized, self-looped, and degree-normalized form of $\bA^{(c)}$, and the product $\tilde{\bA}\bS_t$ is computed over the edge list, so no dense matrix is formed. Here, $\tau$ is the number of propagation steps, and the coefficient $\alpha \in (0,1)$ controls how far the membership mass spreads. The final matrix $\bS_\tau$ defines $\bS^{(c)}$. Equation~\eqref{eq:propagation} is a row-normalized, truncated form of the personalized propagation scheme of~\citet{ppr}. The propagation has no learned parameters, so the decomposition is computed once per graph and reused across training runs.

\subsubsection{Bridge nodes and pooled representations} 

A node whose membership mass in a second child cluster exceeds a threshold is a bridge node. The threshold and the number $k$ of candidates selected per cluster pair are hyperparameters, so the candidate set remains bounded, independent of community size. Each internal cluster also stores two pooled matrices that summarize its child clusters, using the soft assignment as a weight matrix,
\begin{equation}
    \label{eq:pool}
    \xpool^{(c)} = \big(\bS^{(c)}\big)^{\top}\bX^{(c)} \in\mathbb{R}^{K_c\times F},
    \qquad
    \Apool^{(c)} = \big(\bS^{(c)}\big)^{\top}\bA^{(c)}\bS^{(c)} \in\mathbb{R}^{K_c\times K_c}.
\end{equation}
Each row of $\xpool^{(c)}$ is a membership-weighted sum over the nodes of one child cluster. The diagonal entries of $\Apool^{(c)}$ hold the edge mass within each child cluster, and the off-diagonal entries hold the edge mass between pairs of child clusters.

\subsubsection{Edge assignment} 

An edge with both endpoints in the same leaf community ($\ell_u=\ell_v$) is an intra-community edge and is generated by Stage 2. An edge whose endpoints lie in different leaf communities ($\ell_u\neq\ell_v$) is an inter-community edge, owned by the lowest common ancestor of the two communities in $T$, stored under the child that holds each endpoint, and generated by Stage 3. Since every edge falls into exactly one of these two categories, the two edge stages are complementary, and no edge is modeled twice.

\subsection{Stage 1: Node Feature Generation}
\label{sec:method-nodes}

Stage 1 generates node features for each leaf, conditioned on the soft assignment and the corresponding row of $\xpool^{(c)}$, where $c$ is the parent cluster of the leaf. We instantiate the generator as a Transformer~\citep{transformer} over the member nodes of the leaf, where each node attends to the other members and cross-attends to the pooled row. The features of all members of a leaf are therefore generated jointly so that the model can capture correlations among them. We train with a masked reconstruction loss that handles leaves of varying sizes. 

The reconstruction term follows the feature type of the reference graph, detected once from $\bX$. Continuous features are standardized and trained with a Huber loss, and the standardization is inverted at generation. Binary features are trained with binary cross-entropy on the logits, which are then passed through a sigmoid at generation.

\subsection{Stage 2: Intra-Community Edge Generation}
\label{sec:method-intra}

Stage 2 generates the edges within each leaf, which we formulate as an edge prediction problem inside the leaf. Every pair of member nodes is a candidate, the reference edges of the leaf are positives, and the remaining pairs are negatives. A single model is trained over all leaves, which are batched as a disjoint union so that no candidate pair crosses a leaf boundary.

We model each leaf with a variational graph autoencoder~\citep{VGAE}. The encoder first makes the embeddings degree-aware: the feature embedding of a node is scaled and shifted by learned functions of its log-transformed reference degree, following the FiLM conditioning of~\citet{FiLM}, so that each embedding carries the degree budget its node is expected to fill. A graph convolutional network~\citep{GCN} then reads these embeddings together with the leaf adjacency and produces a latent embedding $\bz_i$ per node, which therefore reflects the features of a node's neighbors and not only its own. The decoder scores a pair by a bilinear form $s_{ij}=\bz_i^{\top}\bW\bz_j$, and the edge probability is $\sigma(s_{ij})$, where $\sigma$ is the logistic sigmoid. Training minimizes binary cross-entropy over the observed edges of the leaf and sampled non-edges, with a KL term that keeps the latent embeddings close to a standard normal prior. Both are subsampled, with at most $P$ edges per leaf per epoch and $\nu P$ non-edges, where $\nu$ is the negative sampling ratio, so the per-step cost is set by $P$ and not by the leaf size. 

At generation, there is no reference adjacency to encode, so the latent embeddings are drawn from the prior, and the decoder scores pairs from them together with the degree conditioning. Edges are selected from the score matrix under a per-node degree budget.

\subsection{Stage 3: Inter-Community Edge Generation}
\label{sec:inter}

Stage 3 generates inter-community edges. Scoring every possible pair between two communities is quadratic in community size, so we restrict generation to a candidate pool of fixed size $k$. Candidates are selected from the soft assignments, and bridge nodes are natural candidates by definition.

As in Stage 2, we formulate Stage 3 as an edge prediction task. A pair is scored by a bilinear form over the two node embeddings. We add to this score a cluster-affinity term read from the hierarchy, which measures how often the source's cluster links to the clusters of the destination, weighted by the source's membership in its own cluster. The term is computed from $\bS^{(c)}$ and $\Apool^{(c)}$ alone. It involves no learned parameters, so the model starts from the connectivity already recorded in the hierarchy and only has to adjust it.

\subsection{Training and Generation}
\label{sec:training}

The three stages are trained independently, with no joint objective and no gradient flow between them. This keeps memory bounded by the largest community and avoids balancing a shared loss across three objectives with different scales. 

Generation traverses the tree once. Stage 1 generates the features of every leaf, Stage 2 generates the intra-community edges of each leaf, and Stage 3 visits each internal cluster and generates the edges between its children. Each stage draws its own noise. The noise scale controls how far a stage departs from what it was trained on, and it is the only source of variation between two graphs generated from the same model.

No stage forms a dense $N \times N$ adjacency matrix. The decomposition is linear in the number of edges of each subgraph, and both edge stages train on sampled pairs. One quadratic term remains at generation: Stage 2 scores every pair inside a leaf, which grows with the square of the leaf size. Since the modularity objective places no bound on the leaf size, scalability and memory usage depend on the graph itself. Generation holds at most one batch of clusters in memory at a time.

\section{Experiments and Results}
\label{sec:experiments}

The evaluation covers four aspects. Structural fidelity and memorization test the two requirements stated in Section~\ref{sec:problem}. Downstream utility tests whether the generated graph serves as a substitute for the reference. Scalability measures the cost of generation as the graph grows. 

\subsection{Experimental Setup}
\label{sec:setup}

\subsubsection{Datasets} 

We evaluate on ten attributed graphs in total. Four are used for the full evaluation: Citeseer~\citep{Citeseer} and Cora ML~\citep{CoraML} are citation networks, and Amazon Computers and Amazon Photo~\citep{Amazon} are co-purchase networks. The other six datasets are larger and are used in the main text only to measure scalability: Flickr and Yelp~\citep{SAINT}, Reddit~\citep{SAGE}, DGraphFin~\citep{DGraphFin}, OGBN Products~\citep{OGBN}, and IGB Medium~\citep{IGB}. Dataset details are given in Appendix~\ref{app:datasets}.

\subsubsection{Baselines} 

We compare against four groups of methods. The classical models include the configuration model~\citep{configuration} and DC-SBM~\citep{DC-SBM}. DC-SBM blocks are obtained from a Leiden partition of the reference. The topology-only deep models generate structure without attributes: NetGAN~\citep{NetGAN} and CELL~\citep{CELL}. The attributed models generate structure and node features: GenCAT~\citep{GenCAT}, GraphMaker~\citep{GraphMaker}, and SynGen~\citep{SynGen}. For GraphMaker, we report the asynchronous variant, which the authors identify as the stronger of the two. The subgraph-based models decompose the reference before generating: SaGess~\citep{SaGess} and LGSG~\citep{LGSG}. Further baselines, implementation details, and deviations from the published settings are given in Appendix~\ref{app:baselines}.

\subsubsection{Implementation details}

We implement \method{} in Python with PyTorch~\citep{PyTorch} and PyTorch Geometric~\citep{PyG}. All experiments are conducted on a single machine with an NVIDIA GeForce RTX 3090 GPU (24 GB of memory) and 32 GB of system RAM. The code is available at \url{https://github.com/ahmetTuzen/schema}. For each experiment, we set a 12-hour time limit. Methods that run out of memory (OOM) or time out (TO) are reported as such in the corresponding tables.

\subsubsection{Reporting} 

Unless otherwise stated, each reported number is the mean and standard deviation over 10 independently generated graphs. Due to space limitations, the main text reports the key findings of each experiment. The Appendix provides extended versions of every experiment with per-dataset discussion, a full evaluation on Flickr, and an ablation study (Appendix~\ref{app:ablation}).

\subsection{Structural Fidelity}
\label{sec:structural}

We report four metrics that cover three structural scales: the Wasserstein-1 degree distance and the assortativity difference at the local scale, the triangle count ratio at the higher-order scale, and the ratio of inter-community to intra-community edges at the mesoscopic scale. Ratios are computed relative to the reference and are ideal at one, whereas differences and distances are ideal at zero.

Structural fidelity results are reported in Table~\ref{tab:fidelity}. No model matches the reference at every scale, and the models that lead on one metric are weak on another. Among the learning-based baselines, those that match the reference most closely in any single column either reproduce a large part of it or fail to complete on the larger graphs. \method{} completes on all four datasets. Among the models that generate attributes, it is the closest to the reference Inter/Intra ratio across all datasets. However, the cost is a larger degree distance, so the gain is at the community scale and not at the level of individual nodes.

The triangle ratio should be read together with the memorization results. \method{} scores node pairs independently in both edge stages, and \citet{chanpuriya2021} show that models of this form cannot reach the triangle density of a real graph without reproducing a large fraction of its edges.

\begin{table}[t]
\centering
\caption{Structural fidelity. Results are given as the mean over 10 generated graphs with standard deviation.}
\label{tab:fidelity}
\resizebox{\columnwidth}{!}{
\setlength{\tabcolsep}{2pt}
\begin{tabular}{l | cccc | cccc}
\toprule
\multirow{2}{*}{Model}  & \multicolumn{4}{c}{Citeseer} & \multicolumn{4}{c}{Cora ML} \\
\cmidrule(lr){2-5}\cmidrule(lr){6-9}
        & Deg. W1 & Tri. Rat. & Assort. Diff. & Inter/Intra
        & Deg. W1 & Tri. Rat. & Assort. Diff. & Inter/Intra\\

Config      & $0.007 {\scriptscriptstyle \pm 0.002}$ & $0.026 {\scriptscriptstyle \pm 0.006}$ & $0.059 {\scriptscriptstyle \pm 0.011}$ & $5.207 {\scriptscriptstyle \pm 0.051}$ & $0.047 {\scriptscriptstyle \pm 0.004}$ & $0.097 {\scriptscriptstyle \pm 0.008}$ & $0.048 {\scriptscriptstyle \pm 0.005}$ & $6.653 {\scriptscriptstyle \pm 0.151}$ \\
DC-SBM      & $0.445 {\scriptscriptstyle \pm 0.016}$ & $0.467 {\scriptscriptstyle \pm 0.039}$ & $0.061 {\scriptscriptstyle \pm 0.023}$ & $0.993 {\scriptscriptstyle \pm 0.099}$ & $0.466 {\scriptscriptstyle \pm 0.032}$ & $0.382 {\scriptscriptstyle \pm 0.011}$ & $0.052 {\scriptscriptstyle \pm 0.005}$ & $0.969 {\scriptscriptstyle \pm 0.054}$ \\

GenCAT 		& $0.027 {\scriptscriptstyle \pm 0.004}$ & $0.093 {\scriptscriptstyle \pm 0.011}$ & $0.105 {\scriptscriptstyle \pm 0.006}$ & $3.022 {\scriptscriptstyle \pm 0.194}$ & $0.183 {\scriptscriptstyle \pm 0.007}$ & $0.245 {\scriptscriptstyle \pm 0.014}$ & $0.010 {\scriptscriptstyle \pm 0.007}$ & $1.826 {\scriptscriptstyle \pm 0.177}$ \\
NetGAN 		& $0.612 {\scriptscriptstyle \pm 0.131}$ & $1.024 {\scriptscriptstyle \pm 0.084}$ & $0.063 {\scriptscriptstyle \pm 0.027}$ & $1.472 {\scriptscriptstyle \pm 0.549}$ & $0.875 {\scriptscriptstyle \pm 0.034}$ & $0.489 {\scriptscriptstyle \pm 0.060}$ & $0.012 {\scriptscriptstyle \pm 0.011}$ & $1.810 {\scriptscriptstyle \pm 0.425}$ \\
CELL 		& $0.744 {\scriptscriptstyle \pm 0.012}$ & $0.088 {\scriptscriptstyle \pm 0.008}$ & $0.073 {\scriptscriptstyle \pm 0.015}$ & $3.370 {\scriptscriptstyle \pm 0.071}$ & $0.961 {\scriptscriptstyle \pm 0.030}$ & $0.403 {\scriptscriptstyle \pm 0.016}$ & $0.025 {\scriptscriptstyle \pm 0.004}$ & $2.436 {\scriptscriptstyle \pm 0.108}$ \\

GraphMaker 	& $9.129 {\scriptscriptstyle \pm 0.348}$ & $43.349 {\scriptscriptstyle \pm 3.082}$ & $0.208 {\scriptscriptstyle \pm 0.048}$ & $0.849 {\scriptscriptstyle \pm 0.342}$ & $6.560 {\scriptscriptstyle \pm 0.480}$ & $6.689 {\scriptscriptstyle \pm 0.977}$ & $0.102 {\scriptscriptstyle \pm 0.054}$ & $0.296 {\scriptscriptstyle \pm 0.031}$ \\

SaGess 		& $0.138 {\scriptscriptstyle \pm 0.008}$ & $1.045 {\scriptscriptstyle \pm 0.020}$ & $0.008 {\scriptscriptstyle \pm 0.003}$ & $0.943 {\scriptscriptstyle \pm 0.069}$ & $0.485 {\scriptscriptstyle \pm 0.018}$ & $1.085 {\scriptscriptstyle \pm 0.008}$ & $0.009 {\scriptscriptstyle \pm 0.002}$ & $1.107 {\scriptscriptstyle \pm 0.054}$ \\
LGSG 		& $0.953 {\scriptscriptstyle \pm 0.046}$ & $0.986 {\scriptscriptstyle \pm 0.076}$ & $0.184 {\scriptscriptstyle \pm 0.010}$ & $2.296 {\scriptscriptstyle \pm 0.212}$ & $3.993 {\scriptscriptstyle \pm 0.065}$ & $0.258 {\scriptscriptstyle \pm 0.016}$ & $0.317 {\scriptscriptstyle \pm 0.015}$ & $2.196 {\scriptscriptstyle \pm 0.084}$ \\	

SynGen      & $0.646 {\scriptscriptstyle \pm 0.083}$ & $0.059 {\scriptscriptstyle \pm 0.037}$ & $0.082 {\scriptscriptstyle \pm 0.046}$ & $6.669 {\scriptscriptstyle \pm 1.204}$ & $1.484 {\scriptscriptstyle \pm 0.064}$ & $0.026 {\scriptscriptstyle \pm 0.009}$ & $0.199 {\scriptscriptstyle \pm 0.071}$ & $4.532 {\scriptscriptstyle \pm 0.548}$ \\
 
\textbf{\method} & $0.247 {\scriptscriptstyle \pm 0.015}$ &$0.363 {\scriptscriptstyle \pm 0.021}$  &$0.022 {\scriptscriptstyle \pm 0.005}$ &$0.984 {\scriptscriptstyle \pm 0.090}$ & $1.075 {\scriptscriptstyle \pm 0.024}$ & $0.275 {\scriptscriptstyle \pm 0.019}$ & $0.002 {\scriptscriptstyle \pm 0.002}$ & $1.030 {\scriptscriptstyle \pm 0.083}$\\

\midrule
\multirow{2}{*}{Model} & \multicolumn{4}{c}{Amazon Photo} & \multicolumn{4}{c}{Amazon Computers} \\
\cmidrule(lr){2-5}\cmidrule(lr){6-9}
       & Deg. W1 & Tri. Rat. & Assort. Diff. & Inter/Intra
       & Deg. W1 & Tri. Rat. & Assort. Diff. & Inter/Intra\\

Config & $0.582 {\scriptscriptstyle \pm 0.015}$ & $0.137 {\scriptscriptstyle \pm 0.001}$ & $0.010 {\scriptscriptstyle \pm 0.001}$ & $20.407 {\scriptscriptstyle \pm 1.599}$ & $0.797 {\scriptscriptstyle \pm 0.006}$ & $0.199 {\scriptscriptstyle \pm 0.001}$ & $0.013 {\scriptscriptstyle \pm 0.000}$ & $9.857 {\scriptscriptstyle \pm 0.865}$ \\
DC-SBM & $2.797 {\scriptscriptstyle \pm 0.082}$ & $0.375 {\scriptscriptstyle \pm 0.003}$ & $0.103 {\scriptscriptstyle \pm 0.002}$ & $1.128 {\scriptscriptstyle \pm 0.009}$ & $2.797 {\scriptscriptstyle \pm 0.090}$ & $0.388 {\scriptscriptstyle \pm 0.002}$ & $0.075 {\scriptscriptstyle \pm 0.001}$ & $0.796 {\scriptscriptstyle \pm 0.020}$ \\

GenCAT 		& $1.233 {\scriptscriptstyle \pm 0.046}$ & $0.299 {\scriptscriptstyle \pm 0.008}$ & $0.039 {\scriptscriptstyle \pm 0.003}$ & $1.692 {\scriptscriptstyle \pm 0.261}$ & $1.567 {\scriptscriptstyle \pm 0.044}$ & $0.330 {\scriptscriptstyle \pm 0.009}$ & $0.044 {\scriptscriptstyle \pm 0.003}$ & $0.924 {\scriptscriptstyle \pm 0.102}$ \\

NetGAN 		& $26.537 {\scriptscriptstyle \pm 0.033}$ & $0.877 {\scriptscriptstyle \pm 0.003}$ & $0.680 {\scriptscriptstyle \pm 0.002}$ & $17.509 {\scriptscriptstyle \pm 1.158}$ & \multicolumn{4}{c}{OOM} \\
CELL 	& $2.584 {\scriptscriptstyle \pm 0.055}$ & $0.626 {\scriptscriptstyle \pm 0.004}$ & $0.015 {\scriptscriptstyle \pm 0.002}$ & $1.105 {\scriptscriptstyle \pm 0.020}$ & $1.299 {\scriptscriptstyle \pm 0.023}$ & $0.649 {\scriptscriptstyle \pm 0.003}$ & $0.008 {\scriptscriptstyle \pm 0.001}$ & $1.044 {\scriptscriptstyle \pm 0.066}$ \\

GraphMaker 	& $4.961 {\scriptscriptstyle \pm 0.448}$ & $0.845 {\scriptscriptstyle \pm 0.129}$ & $0.160 {\scriptscriptstyle \pm 0.032}$ & $0.629 {\scriptscriptstyle \pm 0.039}$ & \multicolumn{4}{c}{TO} \\

SaGess  & \multicolumn{4}{c}{TO}  & \multicolumn{4}{c}{TO} \\
LGSG 		& $28.963 {\scriptscriptstyle \pm 0.044}$ & $0.010 {\scriptscriptstyle \pm 0.000}$ & $0.288 {\scriptscriptstyle \pm 0.006}$ & $3.970 {\scriptscriptstyle \pm 0.086}$  & \multicolumn{4}{c}{TO} \\

SynGen & $7.069 {\scriptscriptstyle \pm 1.869}$ & $0.088 {\scriptscriptstyle \pm 0.050}$ & $0.137 {\scriptscriptstyle \pm 0.070}$ & $4.997 {\scriptscriptstyle \pm 1.184}$ & $9.535 {\scriptscriptstyle \pm 2.205}$ & $0.063 {\scriptscriptstyle \pm 0.027}$ & $0.248 {\scriptscriptstyle \pm 0.092}$ & $1.650 {\scriptscriptstyle \pm 0.471}$ \\

\textbf{\method} & $5.417 {\scriptscriptstyle \pm 0.069}$ & $0.294 {\scriptscriptstyle \pm 0.003}$ & $0.024 {\scriptscriptstyle \pm 0.001}$ & $1.159 {\scriptscriptstyle \pm 0.026}$ & $11.084 {\scriptscriptstyle  \pm 0.023}$ & $0.549 {\scriptscriptstyle \pm 0.002}$ & $0.092 {\scriptscriptstyle  \pm 0.001}$ & $0.945 {\scriptscriptstyle \pm 0.085}$\\

\bottomrule
\end{tabular}
}
\end{table}

\subsection{Memorization}
\label{sec:exp-memorization}

A generated graph can score well on every structural metric by reproducing the reference, so fidelity metrics alone cannot separate generalization from memorization. We therefore measure how much of the reference is copied and how much the generated samples differ from one another.

Edge overlap (EO) is the Jaccard similarity between the generated and reference edge sets, self-overlap (SO) is the mean Jaccard similarity between pairs of generated graphs, and neighborhood memorization (NM) is the mean per-node Jaccard similarity between generated and reference neighbor sets. These measures require a node correspondence between the reference and generated graphs. The configuration model and GenCAT are aligned by degree sequence, and the remaining models without a correspondence are excluded. Results are in Table~\ref{tab:memorization}. 

The models with a high triangle ratio in Table~\ref{tab:fidelity} show the highest overlap in Table~\ref{tab:memorization}. Low overlap alone is not evidence of generalization either. The configuration model and GenCAT sit near zero on all three measures, and both distort the community structure.

The results show two memorization patterns. SaGess shows near-complete memorization, with high overlap and high NM, so it reconstructs the reference instead of generating from it. CELL and NetGAN show partial memorization, with moderate overlap and NM. Both stop training once the generated graph reaches a preset overlap with the reference, so the amount they copy is set by this stopping criterion.

\method{} reproduces a small fraction of the reference edges, and its overlap with the reference is at or below its overlap with its own samples on every dataset. The one high value, NM on Citeseer, comes from nodes in leaf communities of at most two nodes, where the reference neighborhood is the only one the model can produce (Appendix~\ref{app:citeseer-nm}).

\begin{table}[t]
\centering
\caption{Memorization. Each result is the mean over 10 generated graphs. NetGAN runs on the largest connected component, not on the full graph.}
\label{tab:memorization}
\resizebox{\columnwidth}{!}{
\footnotesize
\setlength{\tabcolsep}{4pt}
\begin{tabular}{l | ccc | ccc | ccc | ccc}
\toprule
\multirow{2}{*}{Model}  & \multicolumn{3}{c}{Citeseer} & \multicolumn{3}{c}{Cora ML} & \multicolumn{3}{c}{Amazon Photo} & \multicolumn{3}{c}{Amazon Computers}  \\
\cmidrule(lr){2-4} \cmidrule(lr){5-7} \cmidrule(lr){8-10} \cmidrule(lr){11-13}
        & EO & SO & NM 
        & EO & SO & NM 
        & EO & SO & NM 
        & EO & SO & NM \\
Config      & $0.003$ & $0.002$ & $0.001$ & $0.007$ & $0.008$ & $0.003$ & $0.017$ & $0.016$ & $0.007$ & $0.019$ & $0.017$ & $0.006$ \\
DC-SBM      & $0.084$ & $0.063$ & $0.140$ & $0.052$ & $0.046$ & $0.050$ & $0.075$ & $0.066$ & $0.046$ & $0.056$ & $0.051$ & $0.031$ \\

GenCAT & $0.002$ & $0.004$ & $0.001$ & $0.005$ & $0.014$ & $0.002$ & $0.015$ & $0.028$ & $0.007$ & $0.015$ & $0.029$ & $0.005$ \\

NetGAN & $0.228$ & $0.142$ & $0.208$ & $0.174$ & $0.094$ & $0.151$ & $0.027$ & $0.801$ & $0.025$ & \multicolumn{3}{c}{OOM} \\

CELL & $0.384$ & $0.258$ & $0.500$ & $0.442$ & $0.284$ & $0.516$ & $0.337$ & $0.237$ & $0.340$ & $0.348$ & $0.232$ & $0.393$ \\

SaGess & $0.910$ & $0.847$ & $0.904$ & $0.815$ & $0.735$ & $0.795$ & \multicolumn{3}{c}{TO} & \multicolumn{3}{c}{TO} \\

\textbf{\method} & $0.102$ & $0.108$ & $0.254$ & $0.038$ & $0.066$ & $0.068$ & $0.040$ & $0.068$ & $0.035$ & $0.032$ & $0.111$ & $0.024$ \\
\bottomrule

\end{tabular}
}
\end{table}

\subsection{Downstream Utility}
\label{sec:downstream}

A generated graph is useful in practice only if a model trained on it transfers to the reference. We train and test node classifiers on the reference and generated graphs in three settings. TRTR trains and tests on the reference, which serves as the expected ceiling, since a generated graph cannot carry information that the reference does not have. TSTR trains on the generated graph and tests on the reference, and we report its accuracy divided by the TRTR accuracy as utility. TSTS trains and tests on the generated graph, and its difference from TRTR indicates whether the generated task is easier or harder than the reference task. A generator can raise utility by making the task easier, and TSTS detects this. We also compare how TSTS and TRTR rank architectures, using Spearman and Pearson correlations. 

For this task, we report only the models that generate node features. Results are listed in Table~\ref{tab:downstream}. On every dataset where GraphMaker completes, its generated task is easier than the reference, and on Citeseer, its utility exceeds 1. Further analysis in Appendix~\ref{app:homophily} shows that this difference largely comes from inflated homophily.

\method{} retains most of the reference accuracy on every dataset, and its TSTS-TRTR is negative throughout, so its utility does not come from simplifying the task. GenCAT and SynGen also complete on all four datasets but with lower utility. For \method{}, the Pearson correlation is high on every dataset, so the accuracy on the generated graph tracks that of the reference. The Spearman correlation is moderate, so the order of the architectures is only partly preserved.

SynGen retains about a quarter of the reference accuracy across all datasets, which is consistent with its design. It generates structure and features with separate models and joins them with a node-wise aligner, so the dependence between labels and connectivity is not modeled. 

\begin{table}[t]
\caption{Downstream utility. Utility: TSTR to TRTR ratio, where 1 is ideal.}
\centering
\label{tab:downstream}
\resizebox{\columnwidth}{!}{
\setlength{\tabcolsep}{2pt}
\begin{tabular}{l | cccc | cccc}
\toprule
\multirow{2}{*}{Model}  & \multicolumn{4}{c}{Citeseer} & \multicolumn{4}{c}{Cora ML} \\
    \cmidrule(lr){2-5} \cmidrule(lr){6-9}
    & Utility & TSTS-TRTR & Spearman $\rho$ & Pearson $r$
    & Utility & TSTS-TRTR & Spearman $\rho$ & Pearson $r$  \\

GenCAT & $0.752 {\scriptscriptstyle \pm 0.298}$ & $0.017 {\scriptscriptstyle \pm 0.119}$ & $0.261 {\scriptscriptstyle \pm 0.118}$ & $0.682 {\scriptscriptstyle \pm 0.087}$ & $0.686 {\scriptscriptstyle \pm 0.241}$ & $-0.050 {\scriptscriptstyle \pm 0.150}$ & $0.527 {\scriptscriptstyle \pm 0.114}$ & $0.750 {\scriptscriptstyle \pm 0.070}$ \\
GraphMaker & $1.055 {\scriptscriptstyle \pm 0.034}$ & $0.135 {\scriptscriptstyle \pm 0.025}$ & $0.673 {\scriptscriptstyle \pm 0.081}$ & $0.854 {\scriptscriptstyle \pm 0.071}$ & $0.976 {\scriptscriptstyle \pm 0.039}$ & $0.086 {\scriptscriptstyle \pm 0.009}$ & $0.755 {\scriptscriptstyle \pm 0.083}$ & $0.974 {\scriptscriptstyle \pm 0.061}$ \\
SynGen & $0.251 {\scriptscriptstyle \pm 0.064}$ & $-0.503 {\scriptscriptstyle \pm 0.009}$ & $0.050 {\scriptscriptstyle \pm 0.221}$ & $0.176 {\scriptscriptstyle \pm 0.278}$ & $0.263 {\scriptscriptstyle \pm 0.092}$ & $-0.835 {\scriptscriptstyle \pm 0.017}$ & $0.019 {\scriptscriptstyle \pm 0.268}$ & $0.115 {\scriptscriptstyle \pm 0.043}$ \\

\textbf{\method} & $0.861 {\scriptscriptstyle \pm 0.076}$ & $-0.132 {\scriptscriptstyle \pm 0.037}$ & $0.538 {\scriptscriptstyle \pm 0.175}$ & $0.853 {\scriptscriptstyle \pm 0.095}$ & $0.816 {\scriptscriptstyle \pm 0.062}$ & $-0.125 {\scriptscriptstyle \pm 0.029}$ & $0.492 {\scriptscriptstyle \pm 0.107}$ & $0.971 {\scriptscriptstyle \pm 0.019}$ \\
\midrule

\multirow{2}{*}{Model}  & \multicolumn{4}{c}{Amazon Photo} & \multicolumn{4}{c}{Amazon Computers}  \\
    \cmidrule(lr){2-5} \cmidrule(lr){6-9}
    & Utility & TSTS-TRTR & Spearman $\rho$ & Pearson $r$
    & Utility & TSTS-TRTR & Spearman $\rho$ & Pearson $r$  \\
    
GenCAT & $0.470 {\scriptscriptstyle \pm 0.117}$ & $-0.218 {\scriptscriptstyle \pm 0.177}$ & $0.506 {\scriptscriptstyle \pm 0.099}$ & $0.663 {\scriptscriptstyle \pm 0.051}$ & $0.422 {\scriptscriptstyle \pm 0.065}$ & $-0.494 {\scriptscriptstyle \pm 0.063}$ & $0.268 {\scriptscriptstyle \pm 0.275}$ & $0.234 {\scriptscriptstyle \pm 0.229}$ \\
GraphMaker & $0.975 {\scriptscriptstyle \pm 0.011}$ & $0.016 {\scriptscriptstyle \pm 0.007}$ & $0.891 {\scriptscriptstyle \pm 0.030}$ & $0.993 {\scriptscriptstyle \pm 0.009}$ & \multicolumn{4}{c}{TO} \\
SynGen &  $0.220 {\scriptscriptstyle \pm 0.079}$ & $-0.901 {\scriptscriptstyle \pm 0.028}$ & $-0.013 {\scriptscriptstyle \pm 0.278}$ & $-0.251 {\scriptscriptstyle \pm 0.316}$ & $0.248 {\scriptscriptstyle \pm 0.145}$ & $-0.870 {\scriptscriptstyle \pm 0.024}$ & $0.002 {\scriptscriptstyle \pm 0.235}$ & $-0.242 {\scriptscriptstyle \pm 0.372}$ \\

\textbf{\method}  & $0.805 {\scriptscriptstyle \pm 0.100}$ & $-0.111 {\scriptscriptstyle \pm 0.008}$ & $0.443 {\scriptscriptstyle \pm 0.127}$ & $0.994 {\scriptscriptstyle \pm 0.008}$ & $0.771 {\scriptscriptstyle \pm 0.088}$ & $-0.137 {\scriptscriptstyle \pm 0.015}$ & $0.408 {\scriptscriptstyle \pm 0.142}$ & $0.974 {\scriptscriptstyle \pm 0.028}$ \\
\bottomrule

\end{tabular}
}
\end{table}

\subsection{Scalability}
\label{sec:scalability}

\begin{figure}[t]
\centering
\includegraphics[width=\textwidth]{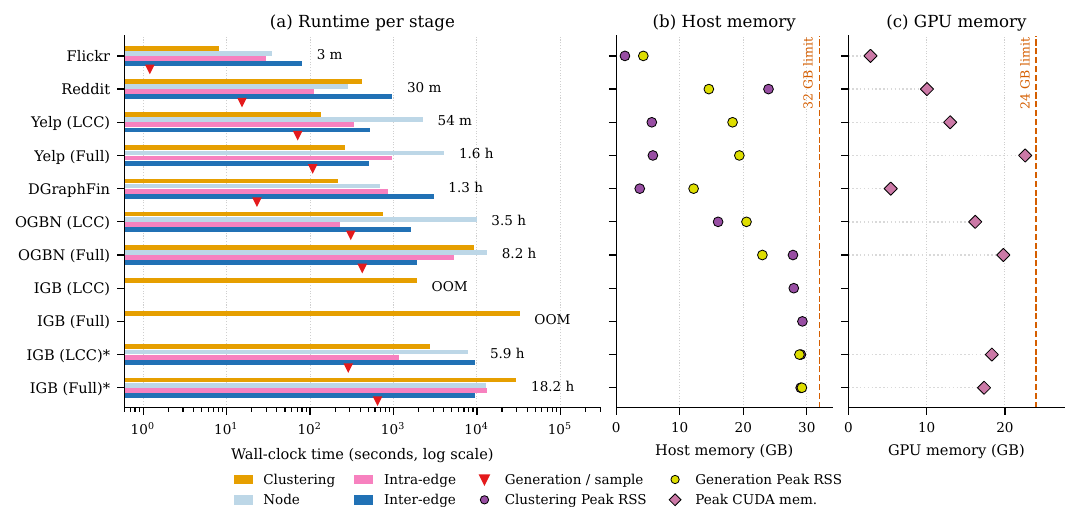}
\caption{Runtime and peak memory usage of \method{} per stage on the six scalability datasets.}
\label{fig:scalability}
\end{figure}

We measure wall-clock time and peak memory per stage, and report training and generation time separately. Peak memory covers both host memory and GPU memory. For graphs that are not fully connected, we run both the full graph and its largest connected component. The batch size is 16. Figure~\ref{fig:scalability} shows the results.

The memory and runtime of clustering and training grow with the number of connected components. Stage 1 dominates training time on larger graphs, whereas on graphs with many inter-community edges, Stage 3 takes the largest share. Generation is inexpensive compared to training.

GPU memory remains bounded by the leaf size and the number of pairs scored per step, as stated in Section~\ref{sec:training}. Since leaf sizes are determined by the modularity objective, peak GPU memory scales with the largest leaf, not with the graph size, so the largest graph does not necessarily use the most GPU memory. IGB Medium does not complete generation under the default settings, since its largest community under the modularity objective has more than 1 million nodes and does not fit in memory. With the leaf size capped at 100{,}000 nodes and a batch size of 8, generation completes, but the cap changes the partition, so this run does not use the same configuration as the other graphs.

\section{Conclusion}

\method{} generates a large attributed graph from a single reference by decomposing it into a hierarchy of soft communities and training three generation stages independently.  Among the models that generate attributes, \method{} is competitive on structural fidelity, keeps memorization low, and retains most of the reference downstream accuracy without making the task easier. Each competitor either memorizes, makes the downstream task easier, fails on larger graphs, or fails to generate features. Generation scales to a graph with 10 million nodes on a single GPU, although this graph requires capping the leaf size. 

The limitations of \method{} lead to several directions for future work. Both edge stages generate edges from pairwise scores, so at low overlap, the generated graphs have fewer triangles than the reference. One direction is to condition the edge generators on triangle counts to better match this statistic. Another direction is to extend the framework to other graph types, such as heterogeneous graphs and dynamic graphs.

\section*{Acknowledgment}

This paper has been funded by the SFI NorwAI, (Centre for Research-based Innovation, 309834). The authors gratefully acknowledge the financial support from the Research Council of Norway and the partners of the SFI NorwAI.

\bibliography{references}
\bibliographystyle{iclr2027_conference}

\clearpage

\appendix

\section{Appendix}

\subsection{Datasets}
\label{app:datasets}

The four evaluation graphs (Citeseer, Cora ML, Amazon Photo, and Amazon Computers) have the same format. Nodes carry binary bag-of-words vectors, edges are undirected, and every node has exactly one class label. In the two citation networks, nodes are papers, edges are citations, features indicate whether a dictionary word occurs in the paper, and classes are research areas. Amazon Photo and Amazon Computers are co-purchase networks in which nodes are products, edges connect frequently co-purchased products, features are derived from product reviews, and product categories serve as labels. Citeseer and Cora ML have higher-dimensional sparse features, while the Amazon graphs are denser in both attributes and degree.
 
The six scalability graphs are larger and carry continuous features, which are dense on every graph except Flickr. Flickr is an image network in which nodes are images, edges connect images that share properties such as a common gallery or location, features are built from words that describe the images, and node labels are image categories. Reddit is a social network where nodes are posts, an edge means the same user commented on both posts, features are built from posts and entries, and labels are subreddits. Yelp is a social network in which nodes are users and edges are friendships, features are embeddings of each user's reviews, and labels are the types of businesses the user reviewed. It is the only multi-label graph. OGBN Products is a co-purchase network, with features derived from product descriptions and product categories as labels. DGraphFin is a financial network in which nodes are users of a lending platform, and an edge means one user listed another as an emergency contact. Its features are anonymized profile fields, and its four label values cover two foreground classes, normal and fraudulent, plus two background classes for unlabeled users. IGB Medium is a citation network in which features are language model embeddings of paper abstracts.

Detailed dataset statistics are given in Table~\ref{tab:datasets}. Graphs with many connected components are harder for the hierarchy: modularity gains nothing by merging nodes across components, so each component yields at least one community. Citeseer is the extreme case in relative terms: only 63.7\% of its nodes lie in the largest component, and 16.4\% lie in components of at most two nodes.

\begin{table}[b]
\centering
\caption{Dataset statistics. \# Con. Comp. is the number of connected components. LCC is the fraction of nodes in the largest connected component. $F$ is the number of node attributes. Feat.\ den.\ is the fraction of non-zero entries in the attribute matrix. $C$ is the number of classes.}
\label{tab:datasets}
\resizebox{\columnwidth}{!}{
\begin{tabular}{lrrrrcccc}
\toprule
Dataset  & $N$ & $|E|$ & Avg.\ deg. & \# Con. Comp. & LCC & $F$ & Feat. den. & $C$ \\
\midrule

Citeseer             & 3{,}327        & 4{,}552        & 2.74   & 438        & 63.7\% & 3{,}703      & 0.009 & 6 \\
Cora ML              & 2{,}995        & 8{,}158        & 5.45   & 61         & 93.8\% & 2{,}879      & 0.018 & 7 \\
Amazon Photo         & 7{,}650        & 119{,}081      & 31.13  & 136        & 97.9\% & 745          & 0.347 & 8 \\
Amazon Computers     & 13{,}752       & 245{,}861      & 35.76  & 314        & 97.3\% & 767          & 0.348 & 10 \\
Flickr               & 89{,}250       & 449{,}878      & 10.08  & 1          & 100\%  & 500          & 0.464 & 7 \\
Reddit               & 232{,}965      & 57{,}307{,}946 & 491.99 & 1          & 100\%  & 602          & 1.000 & 41 \\
Yelp                 & 716{,}847      & 6{,}618{,}986  & 18.47  & 6{,}288    & 98.2\% & 300          & 1.000 & 100 \\
OGBN Products        & 2{,}449{,}029  & 61{,}859{,}012 & 50.52  & 52{,}658   & 97.4\% & 100          & 0.990 & 47 \\
DGraphFin            & 3{,}700{,}550  & 3{,}997{,}260  & 1.08   & 1          & 100\%  & 17           & 0.942 & 4 \\
IGB Medium           & 10{,}000{,}000 & 59{,}997{,}454 & 12.00  & 111{,}011  & 98.8\% & 128$^\dagger$     & 1.000 & - \\
\bottomrule
\end{tabular}
}
\begin{flushleft}\footnotesize
$^\dagger$ The original graph has 1{,}024 features, of which we keep the first 128 to fit in memory.
\end{flushleft}
\end{table}

\subsection{Baselines}
\label{app:baselines}

\subsubsection{Classical models}

We implement six random graph models, two of which appear in the main text, and two graph autoencoders adapted for generation. None of them generates node attributes. The random graph models define a distribution over edge sets only, and the autoencoders read node features as encoder input but decode only edges. We therefore use them only in the structural and memorization experiments, as structural baselines and as a check in the evaluation. 

We generate the Erd\H{o}s--R\'enyi (ER), Watts--Strogatz (WS), Barab\'asi--Albert (BA), Chung--Lu (CL), and configuration model (Config) with the corresponding NetworkX generators \citep{NetworkX}, two of which use the linear-time samplers of \citet{BatageljBrandes} for ER $G(n,p)$ and \citet{MillerHagberg} for CL instead of iterating over all pairs. The degree-corrected stochastic block model (DC-SBM) is implemented in NumPy \citep{NumPy}. Every model is fitted to the same reference graph as the other baselines, and all of them operate on the simple undirected edge list with self-loops and duplicate edges removed. Each model generates 10 samples per dataset with independent seeds, and the same 10 seeds are used across models and datasets.

The models differ in how much information they receive from the reference. In order of increasing information: ER $G(n,p)$ receives the node count and the edge density $p = M / \binom{N}{2}$, where $M = |E|$ is the reference edge count. ER $G(n,m)$ receives the node count and the exact edge count. WS receives the node count and the mean degree $k$, rounded to the nearest integer, and its rewiring probability is not fitted but fixed at $\beta = 0.1$. BA receives the node count and the per-node attachment count $m$, which we set to the reference edge count divided by the node count, rounded to the nearest integer.

CL receives the full reference degree sequence and treats it as a sequence of expected degrees, so the generated degrees match it only in expectation. Config receives the same sequence and matches it exactly, but samples a multi-graph. Reducing parallel edges to a single edge and removing self-loops lowers the realized degrees.  

Our DC-SBM implementation follows~\citet{DC-SBM}. The model is a Poisson multi-graph in which the expected number of edges between nodes $i$ and $j$ is $\theta_i \theta_j \omega_{b_i b_j}$, where $b_i$ is the block of node $i$, $\omega_{rs}$ is the edge budget between blocks $r$ and $s$, and $\theta_i = d_i / \sum_{j: b_j = b_i} d_j$ is a within-block degree parameter, with $d_i$ the degree of node $i$. Under the maximum-likelihood fit for known blocks, $\omega_{rs}$ is the observed edge count $m_{rs}$ between blocks $r$ and $s$, and twice the internal edge count when $r = s$.

The $\theta_i$ sum to one within each block, so the per-pair Poisson means add up to $\omega_{rs}$: the total number of edges between two blocks is itself Poisson with that mean. Conditional on that total, the two endpoints of each edge are independent draws proportional to $\theta$ within their blocks. We therefore sample one Poisson count per occupied block pair and then place the endpoints, which costs $O(M)$ instead of the $O(N^2)$ of a per-pair formulation. This sampler is exact, not an approximation. The sampled multi-graph is projected to a simple undirected graph in the same way as the configuration model.

The blocks come from a Leiden partition \citep{Leiden} of the reference at resolution $1.0$, not from the node labels, so the model receives no label information. The same Leiden setting defines the reference communities in the mesoscopic metrics, so DC-SBM's mesoscopic scores measure how well the Poisson sampler preserves the given partition, not whether it can find community structure. \method{} builds its leaves from the same partition. DC-SBM therefore shares \method{}'s communities and replaces the learned edge model and the hierarchy with a Poisson block model.

\textbf{Generated edge counts.} Four of the classical models systematically miss the reference edge count, and ER $G(n,p)$ and CL match it only in expectation. BA and WS quantize the target density to an integer parameter, so their realized edge counts are $m(N-m)$ and $\lfloor k/2 \rfloor N$. Config loses edges when parallel edges and self-loops are removed. DC-SBM loses edges when the Poisson multi-graph is projected onto a simple graph. We do not correct any of these, and the Deg.\ Rat.\ column of the structural fidelity tables shows the resulting deviation.

\subsubsection{Graph autoencoders}

We implement the graph autoencoder (GAE) and the variational graph autoencoder (VGAE) \citep{VGAE} as two-layer graph convolutional encoders with an inner-product decoder, trained on positive edges and an equal number of negative edges. Both read node features as encoder input and decode only edges. We change the standard setup in two ways.

First, we divide the KL term by the number of nodes. Without this normalization, the KL term dominates the objective and the posterior collapses to the prior, so the latents carry almost no information about the graph.

Second, the decoder scores every node pair, so it needs a rule for turning scores into edges. We give it $N$ and the reference edge count $M$, as for ER $G(n,m)$, and keep the $M$ highest-scoring pairs after a Gumbel perturbation at temperature $0.2$. The edge count is then fixed, and the node that receives those edges is determined by the decoder.

Since GAE defines no prior, its only valid generation setting is reconstruction from the encoded reference. VGAE supports both reconstruction and prior sampling, but we report only the prior version.

\subsubsection{Published implementations}

For the remaining baselines, we use the authors' released code with the hyperparameters reported in their papers or source code. The notes below list where we deviate from the published setting, or where the released code forces a choice, because each affects how the corresponding results should be interpreted.

\begin{itemize}
\item \textbf{CELL and NetGAN.} Both stop training when the generated graph reaches a preset edge overlap with the reference, so their edge overlap is fixed in advance. We use the published stopping criteria on every dataset. Their memorization results, therefore, reflect the stopping rule rather than generalization.

\item \textbf{NetGAN.} NetGAN works on the largest connected component, so its generated graphs cover only the nodes in that component. All NetGAN results are computed against the largest connected component of the reference. Size-dependent quantities such as the triangle count and the Jaccard denominators are therefore taken over a smaller graph than for the other rows.

\item \textbf{GenCAT.} GenCAT takes a class mixing matrix and class-wise attribute statistics as input, so it receives more information from the reference than most baselines. It also takes the reference degree sequence as input, which keeps its degree distance low. Its generated features are continuous and nearly uniformly dense, so feature-marginal comparisons against the binary references are not meaningful.

\item \textbf{GraphMaker.} The model is label-conditional, so like GenCAT, it receives label information. The released implementation stores a dense $N \times N$ adjacency matrix, which limits the size of the graph it can handle. The main text reports the asynchronous variant, and the appendix reports both.

\item \textbf{SaGess.} We run SaGess as published, with no deviations.

\item \textbf{SynGen.} The model is unnamed in the paper; we use the name from the repository. We use the KDE feature generator, the tool's default, which has the highest feature correlation in the authors' ablation study. Two properties of the method directly affect the generated graphs. First, SynGen's R-MAT \citep{RMAT} structural generator uses $\lceil \log_2 N \rceil$ bits per node identifier, so the node count rises to $2^{\lceil \log_2 N \rceil}$ minus any unused identifiers. The edge count is matched exactly, so the generated graphs are sparser than the reference. Second, structure and features are generated independently and joined by a node-wise aligner, which is trained to predict a node's feature row from single-node structural summaries. The aligner, therefore, cannot represent homophily.

\item \textbf{LGSG.} The model generates many small subgraphs with node embeddings and assembles them into one graph. The authors' threshold-based assembler produces graphs with varying sizes across datasets, so we use the node assembler, which targets the reference node count. The target is not always met, and the edge count is not controlled. Each of the 10 generated graphs is assembled from a disjoint slice of the generated subgraphs.

\item \textbf{SANGEA.} No implementation is publicly available, so we discuss it in the related work and do not include it in the experiments.

\item \textbf{CGT.} The model learns and generates a separate computation graph per node and does not generate a single graph as the other models do. None of the structural or attribute metrics is defined on its output, so we exclude it.

\end{itemize}

\subsection{Full Structural Fidelity}
\label{app:fidelity}

\begin{table}[t]
\centering
\caption{Citeseer structural fidelity results.}
\label{tab:citeseer}
\resizebox{\columnwidth}{!}{
\setlength{\tabcolsep}{2pt}
\begin{tabular}{l|c|c|c|c|c|c|c|c}
\toprule
Model  & Deg. Rat. & Deg. W1 & Tri. Rat. & Clus. C. Rat. & Assort. Diff. & Mod. Diff. & Inter/Intra & CPL Rat.\\
\midrule
ER $G(n,p)$ &$0.992 {\scriptscriptstyle \pm 0.013}$ &$0.834 {\scriptscriptstyle \pm 0.018}$ &$0.004 {\scriptscriptstyle \pm 0.001}$ &$0.008 {\scriptscriptstyle \pm 0.003}$ &$0.045 {\scriptscriptstyle \pm 0.022}$ &$0.194 {\scriptscriptstyle \pm 0.004}$ &$6.404 {\scriptscriptstyle \pm 0.137}$ &$0.852 {\scriptscriptstyle \pm 0.010}$ \\
ER $G(n,m)$ &$1.000 {\scriptscriptstyle \pm 0.000}$ &$0.846 {\scriptscriptstyle \pm 0.016}$ &$0.004 {\scriptscriptstyle \pm 0.002}$ &$0.008 {\scriptscriptstyle \pm 0.005}$ &$0.041 {\scriptscriptstyle \pm 0.007}$ &$0.195 {\scriptscriptstyle \pm 0.003}$ &$6.409 {\scriptscriptstyle \pm 0.102}$ &$0.851 {\scriptscriptstyle \pm 0.004}$ \\
WS &$0.731 {\scriptscriptstyle \pm 0.000}$ &$1.415 {\scriptscriptstyle \pm 0.012}$ &$0.000 {\scriptscriptstyle \pm 0.000}$ &$0.001 {\scriptscriptstyle \pm 0.002}$ &$0.091 {\scriptscriptstyle \pm 0.021}$ &$0.072 {\scriptscriptstyle \pm 0.001}$ &$0.298 {\scriptscriptstyle \pm 0.012}$ &$14.745 {\scriptscriptstyle \pm 5.090}$ \\
BA &$0.731 {\scriptscriptstyle \pm 0.000}$ &$0.852 {\scriptscriptstyle \pm 0.026}$ &$0.000 {\scriptscriptstyle \pm 0.000}$ &$0.000 {\scriptscriptstyle \pm 0.000}$ &$0.137 {\scriptscriptstyle \pm 0.015}$ &$0.065 {\scriptscriptstyle \pm 0.001}$ &$0.344 {\scriptscriptstyle \pm 0.017}$ &$0.886 {\scriptscriptstyle \pm 0.053}$ \\
CL &$0.997 {\scriptscriptstyle \pm 0.010}$ &$0.441 {\scriptscriptstyle \pm 0.018}$ &$0.040 {\scriptscriptstyle \pm 0.008}$ &$0.035 {\scriptscriptstyle \pm 0.006}$ &$0.057 {\scriptscriptstyle \pm 0.006}$ &$0.285 {\scriptscriptstyle \pm 0.004}$ &$7.939 {\scriptscriptstyle \pm 0.153}$ &$0.577 {\scriptscriptstyle \pm 0.007}$ \\
Config &$0.997 {\scriptscriptstyle \pm 0.001}$ &$0.007 {\scriptscriptstyle \pm 0.002}$ &$0.026 {\scriptscriptstyle \pm 0.006}$ &$0.027 {\scriptscriptstyle \pm 0.006}$ &$0.059 {\scriptscriptstyle \pm 0.011}$ &$0.185 {\scriptscriptstyle \pm 0.002}$ &$5.207 {\scriptscriptstyle \pm 0.051}$ &$0.654 {\scriptscriptstyle \pm 0.004}$ \\
DC-SBM &$0.854 {\scriptscriptstyle \pm 0.012}$ &$0.445 {\scriptscriptstyle \pm 0.016}$ &$0.467 {\scriptscriptstyle \pm 0.039}$ &$0.576 {\scriptscriptstyle \pm 0.042}$ &$0.061 {\scriptscriptstyle \pm 0.023}$ &$0.008 {\scriptscriptstyle \pm 0.004}$ &$0.993 {\scriptscriptstyle \pm 0.099}$ &$0.815 {\scriptscriptstyle \pm 0.032}$ \\
GenCAT &$0.990 {\scriptscriptstyle \pm 0.001}$ &$0.027 {\scriptscriptstyle \pm 0.004}$ &$0.093 {\scriptscriptstyle \pm 0.011}$ &$0.097 {\scriptscriptstyle \pm 0.011}$ &$0.105 {\scriptscriptstyle \pm 0.006}$ &$0.124 {\scriptscriptstyle \pm 0.007}$ &$3.022 {\scriptscriptstyle \pm 0.194}$ &$0.737 {\scriptscriptstyle \pm 0.009}$ \\
GAE &$1.000 {\scriptscriptstyle \pm 0.000}$ &$2.156 {\scriptscriptstyle \pm 0.007}$ &$11.630 {\scriptscriptstyle \pm 0.062}$ &$2.357 {\scriptscriptstyle \pm 0.015}$ &$0.181 {\scriptscriptstyle \pm 0.003}$ &$0.075 {\scriptscriptstyle \pm 0.002}$ &$0.909 {\scriptscriptstyle \pm 0.058}$ &$0.740 {\scriptscriptstyle \pm 0.041}$ \\
VGAE &$1.000 {\scriptscriptstyle \pm 0.000}$ &$0.969 {\scriptscriptstyle \pm 0.019}$ &$0.528 {\scriptscriptstyle \pm 0.018}$ &$0.407 {\scriptscriptstyle \pm 0.014}$ &$0.179 {\scriptscriptstyle \pm 0.012}$ &$0.305 {\scriptscriptstyle \pm 0.006}$ &$9.932 {\scriptscriptstyle \pm 0.304}$ &$0.532 {\scriptscriptstyle \pm 0.005}$ \\
NetGAN &$0.850 {\scriptscriptstyle \pm 0.000}$ &$0.612 {\scriptscriptstyle \pm 0.131}$ &$1.024 {\scriptscriptstyle \pm 0.084}$ &$1.279 {\scriptscriptstyle \pm 0.276}$ &$0.063 {\scriptscriptstyle \pm 0.027}$ &$0.038 {\scriptscriptstyle \pm 0.039}$ &$1.472 {\scriptscriptstyle \pm 0.549}$ &$0.679 {\scriptscriptstyle \pm 0.111}$ \\
CELL &$1.000 {\scriptscriptstyle \pm 0.000}$ &$0.744 {\scriptscriptstyle \pm 0.012}$ &$0.088 {\scriptscriptstyle \pm 0.008}$ &$0.184 {\scriptscriptstyle \pm 0.018}$ &$0.073 {\scriptscriptstyle \pm 0.015}$ &$0.118 {\scriptscriptstyle \pm 0.003}$ &$3.370 {\scriptscriptstyle \pm 0.071}$ &$0.946 {\scriptscriptstyle \pm 0.006}$ \\
GraphMaker S &$0.864 {\scriptscriptstyle \pm 0.036}$ &$0.464 {\scriptscriptstyle \pm 0.047}$ &$0.036 {\scriptscriptstyle \pm 0.020}$ &$0.072 {\scriptscriptstyle \pm 0.031}$ &$0.180 {\scriptscriptstyle \pm 0.071}$ &$0.146 {\scriptscriptstyle \pm 0.011}$ &$3.717 {\scriptscriptstyle \pm 0.370}$ &$0.830 {\scriptscriptstyle \pm 0.021}$ \\
GraphMaker A &$4.284 {\scriptscriptstyle \pm 0.125}$ &$9.129 {\scriptscriptstyle \pm 0.348}$ &$43.349 {\scriptscriptstyle \pm 3.082}$ &$2.167 {\scriptscriptstyle \pm 0.071}$ &$0.208 {\scriptscriptstyle \pm 0.048}$ &$0.131 {\scriptscriptstyle \pm 0.005}$ &$0.849 {\scriptscriptstyle \pm 0.342}$ &$0.570 {\scriptscriptstyle \pm 0.010}$ \\
SaGess &$0.995 {\scriptscriptstyle \pm 0.001}$ &$0.138 {\scriptscriptstyle \pm 0.008}$ &$1.045 {\scriptscriptstyle \pm 0.020}$ &$0.986 {\scriptscriptstyle \pm 0.015}$ &$0.008 {\scriptscriptstyle \pm 0.003}$ &$0.006 {\scriptscriptstyle \pm 0.001}$ &$0.943 {\scriptscriptstyle \pm 0.069}$ &$0.901 {\scriptscriptstyle \pm 0.014}$ \\
LGSG &$1.113 {\scriptscriptstyle \pm 0.015}$ &$0.953 {\scriptscriptstyle \pm 0.046}$ &$0.986 {\scriptscriptstyle \pm 0.076}$ &$0.044 {\scriptscriptstyle \pm 0.005}$ &$0.184 {\scriptscriptstyle \pm 0.010}$ &$0.114 {\scriptscriptstyle \pm 0.013}$ &$2.296 {\scriptscriptstyle \pm 0.212}$ &$0.450 {\scriptscriptstyle \pm 0.018}$ \\
SynGen &$0.813 {\scriptscriptstyle \pm 0.001}$ &$0.646 {\scriptscriptstyle \pm 0.083}$ &$0.059 {\scriptscriptstyle \pm 0.037}$ &$0.065 {\scriptscriptstyle \pm 0.033}$ &$0.082 {\scriptscriptstyle \pm 0.046}$ &$0.247 {\scriptscriptstyle \pm 0.048}$ &$6.669 {\scriptscriptstyle \pm 1.204}$ &$0.668 {\scriptscriptstyle \pm 0.096}$ \\
\midrule
\textbf{\method} &$0.996 {\scriptscriptstyle \pm 0.005}$ &$0.247 {\scriptscriptstyle \pm 0.015}$ &$0.363 {\scriptscriptstyle \pm 0.021}$ &$0.368 {\scriptscriptstyle \pm 0.020}$ &$0.022 {\scriptscriptstyle \pm 0.005}$ &$0.018 {\scriptscriptstyle \pm 0.001}$ &$0.984 {\scriptscriptstyle \pm 0.090}$ &$0.884 {\scriptscriptstyle \pm 0.015}$ \\
		
\bottomrule
\end{tabular}
}
\end{table}

\begin{table}[t]
\centering
\caption{Cora ML structural fidelity results.}
\label{tab:coraml}
\resizebox{\columnwidth}{!}{
\setlength{\tabcolsep}{2pt}
\begin{tabular}{l|c|c|c|c|c|c|c|c}
\toprule
Model  & Deg. Rat. & Deg. W1 & Tri. Rat. & Clus. C. Rat. & Assort. Diff. & Mod. Diff. & Inter/Intra & CPL Rat.\\
\midrule
ER $G(n,p)$ & $0.994 {\scriptscriptstyle \pm 0.010}$ & $2.448 {\scriptscriptstyle \pm 0.031}$ & $0.004 {\scriptscriptstyle \pm 0.001}$ & $0.014 {\scriptscriptstyle \pm 0.003}$ & $0.067 {\scriptscriptstyle \pm 0.010}$ & $0.335 {\scriptscriptstyle \pm 0.003}$ & $7.241 {\scriptscriptstyle \pm 0.086}$ & $0.934 {\scriptscriptstyle \pm 0.005}$ \\
ER $G(n,m)$ & $1.000 {\scriptscriptstyle \pm 0.000}$ & $2.459 {\scriptscriptstyle \pm 0.026}$ & $0.005 {\scriptscriptstyle \pm 0.001}$ & $0.015 {\scriptscriptstyle \pm 0.002}$ & $0.074 {\scriptscriptstyle \pm 0.012}$ & $0.336 {\scriptscriptstyle \pm 0.001}$ & $7.245 {\scriptscriptstyle \pm 0.106}$ & $0.932 {\scriptscriptstyle \pm 0.001}$ \\
WS & $0.734 {\scriptscriptstyle \pm 0.000}$ & $3.278 {\scriptscriptstyle \pm 0.011}$ & $0.413 {\scriptscriptstyle \pm 0.009}$ & $3.076 {\scriptscriptstyle \pm 0.072}$ & $0.049 {\scriptscriptstyle \pm 0.019}$ & $0.099 {\scriptscriptstyle \pm 0.005}$ & $0.800 {\scriptscriptstyle \pm 0.043}$ & $2.008 {\scriptscriptstyle \pm 0.045}$ \\
BA & $1.100 {\scriptscriptstyle \pm 0.000}$ & $1.078 {\scriptscriptstyle \pm 0.052}$ & $0.068 {\scriptscriptstyle \pm 0.006}$ & $0.067 {\scriptscriptstyle \pm 0.005}$ & $0.018 {\scriptscriptstyle \pm 0.008}$ & $0.364 {\scriptscriptstyle \pm 0.002}$ & $7.278 {\scriptscriptstyle \pm 0.206}$ & $0.734 {\scriptscriptstyle \pm 0.002}$ \\
CL & $1.002 {\scriptscriptstyle \pm 0.015}$ & $0.451 {\scriptscriptstyle \pm 0.064}$ & $0.147 {\scriptscriptstyle \pm 0.009}$ & $0.141 {\scriptscriptstyle \pm 0.005}$ & $0.053 {\scriptscriptstyle \pm 0.005}$ & $0.379 {\scriptscriptstyle \pm 0.005}$ & $7.431 {\scriptscriptstyle \pm 0.141}$ & $0.750 {\scriptscriptstyle \pm 0.006}$ \\
Config & $0.991 {\scriptscriptstyle \pm 0.001}$ & $0.047 {\scriptscriptstyle \pm 0.004}$ & $0.097 {\scriptscriptstyle \pm 0.008}$ & $0.104 {\scriptscriptstyle \pm 0.009}$ & $0.048 {\scriptscriptstyle \pm 0.005}$ & $0.342 {\scriptscriptstyle \pm 0.002}$ & $6.653 {\scriptscriptstyle \pm 0.151}$ & $0.790 {\scriptscriptstyle \pm 0.003}$ \\
DC-SBM & $0.925 {\scriptscriptstyle \pm 0.010}$ & $0.466 {\scriptscriptstyle \pm 0.032}$ & $0.382 {\scriptscriptstyle \pm 0.011}$ & $0.524 {\scriptscriptstyle \pm 0.012}$ & $0.052 {\scriptscriptstyle \pm 0.005}$ & $0.004 {\scriptscriptstyle \pm 0.003}$ & $0.969 {\scriptscriptstyle \pm 0.054}$ & $0.910 {\scriptscriptstyle \pm 0.007}$ \\
GenCAT & $0.966 {\scriptscriptstyle \pm 0.001}$ & $0.183 {\scriptscriptstyle \pm 0.007}$ & $0.245 {\scriptscriptstyle \pm 0.014}$ & $0.349 {\scriptscriptstyle \pm 0.019}$ & $0.010 {\scriptscriptstyle \pm 0.007}$ & $0.130 {\scriptscriptstyle \pm 0.010}$ & $1.826 {\scriptscriptstyle \pm 0.177}$ & $0.911 {\scriptscriptstyle \pm 0.004}$ \\
GAE & $1.000 {\scriptscriptstyle \pm 0.000}$ & $3.361 {\scriptscriptstyle \pm 0.016}$ & $6.098 {\scriptscriptstyle \pm 0.038}$ & $2.092 {\scriptscriptstyle \pm 0.018}$ & $0.220 {\scriptscriptstyle \pm 0.002}$ & $0.015 {\scriptscriptstyle \pm 0.002}$ & $0.517 {\scriptscriptstyle \pm 0.074}$ & $0.828 {\scriptscriptstyle \pm 0.010}$ \\
VGAE & $1.000 {\scriptscriptstyle \pm 0.000}$ & $1.169 {\scriptscriptstyle \pm 0.040}$ & $0.399 {\scriptscriptstyle \pm 0.010}$ & $0.546 {\scriptscriptstyle \pm 0.010}$ & $0.057 {\scriptscriptstyle \pm 0.016}$ & $0.319 {\scriptscriptstyle \pm 0.002}$ & $6.275 {\scriptscriptstyle \pm 0.140}$ & $0.780 {\scriptscriptstyle \pm 0.003}$ \\
NetGAN & $0.850 {\scriptscriptstyle \pm 0.000}$ & $0.875 {\scriptscriptstyle \pm 0.034}$ & $0.489 {\scriptscriptstyle \pm 0.060}$ & $0.634 {\scriptscriptstyle \pm 0.118}$ & $0.012 {\scriptscriptstyle \pm 0.011}$ & $0.088 {\scriptscriptstyle \pm 0.037}$ & $1.810 {\scriptscriptstyle \pm 0.425}$ & $0.850 {\scriptscriptstyle \pm 0.049}$ \\
CELL & $1.000 {\scriptscriptstyle \pm 0.000}$ & $0.961 {\scriptscriptstyle \pm 0.030}$ & $0.403 {\scriptscriptstyle \pm 0.016}$ & $0.643 {\scriptscriptstyle \pm 0.024}$ & $0.025 {\scriptscriptstyle \pm 0.004}$ & $0.115 {\scriptscriptstyle \pm 0.005}$ & $2.436 {\scriptscriptstyle \pm 0.108}$ & $0.933 {\scriptscriptstyle \pm 0.004}$ \\
GraphMaker S & $0.640 {\scriptscriptstyle \pm 0.027}$ & $1.961 {\scriptscriptstyle \pm 0.145}$ & $0.021 {\scriptscriptstyle \pm 0.006}$ & $0.116 {\scriptscriptstyle \pm 0.022}$ & $0.203 {\scriptscriptstyle \pm 0.062}$ & $0.104 {\scriptscriptstyle \pm 0.006}$ & $2.119 {\scriptscriptstyle \pm 0.111}$ & $1.160 {\scriptscriptstyle \pm 0.031}$ \\
GraphMaker A & $2.170 {\scriptscriptstyle \pm 0.089}$ & $6.560 {\scriptscriptstyle \pm 0.480}$ & $6.689 {\scriptscriptstyle \pm 0.977}$ & $1.917 {\scriptscriptstyle \pm 0.131}$ & $0.102 {\scriptscriptstyle \pm 0.054}$ & $0.025 {\scriptscriptstyle \pm 0.005}$ & $0.296 {\scriptscriptstyle \pm 0.031}$ & $0.882 {\scriptscriptstyle \pm 0.007}$ \\
SaGess & $0.994 {\scriptscriptstyle \pm 0.001}$ & $0.485 {\scriptscriptstyle \pm 0.018}$ & $1.085 {\scriptscriptstyle \pm 0.008}$ & $0.983 {\scriptscriptstyle \pm 0.018}$ & $0.009 {\scriptscriptstyle \pm 0.002}$ & $0.017 {\scriptscriptstyle \pm 0.003}$ & $1.107 {\scriptscriptstyle \pm 0.054}$ & $0.939 {\scriptscriptstyle \pm 0.011}$ \\
LGSG & $0.536 {\scriptscriptstyle \pm 0.009}$ & $3.993 {\scriptscriptstyle \pm 0.065}$ & $0.258 {\scriptscriptstyle \pm 0.016}$ & $0.012 {\scriptscriptstyle \pm 0.001}$ & $0.317 {\scriptscriptstyle \pm 0.015}$ & $0.250 {\scriptscriptstyle \pm 0.017}$ & $2.196 {\scriptscriptstyle \pm 0.084}$ & $0.455 {\scriptscriptstyle \pm 0.008}$ \\
SynGen & $0.731 {\scriptscriptstyle \pm 0.000}$ & $1.484 {\scriptscriptstyle \pm 0.064}$ & $0.026 {\scriptscriptstyle \pm 0.009}$ & $0.077 {\scriptscriptstyle \pm 0.023}$ & $0.199 {\scriptscriptstyle \pm 0.071}$ & $0.235 {\scriptscriptstyle \pm 0.013}$ & $4.532 {\scriptscriptstyle \pm 0.548}$ & $1.061 {\scriptscriptstyle \pm 0.025}$ \\
\midrule
\textbf{\method} & $1.032 {\scriptscriptstyle \pm 0.004}$ & $1.075 {\scriptscriptstyle \pm 0.024}$ & $0.275 {\scriptscriptstyle \pm 0.019}$ & $0.202 {\scriptscriptstyle \pm 0.013}$ & $0.002 {\scriptscriptstyle \pm 0.002}$ & $0.042 {\scriptscriptstyle \pm 0.003}$ & $1.030 {\scriptscriptstyle \pm 0.083}$ & $0.845 {\scriptscriptstyle \pm 0.011}$ \\

\bottomrule
\end{tabular}
}
\end{table}

Structural fidelity metrics follow three conventions. Ratios divide the value of a metric on the generated graph $\Gen$ by its value on the reference graph $G$. Differences are absolute, $|m(G) - m(\Gen)|$ for a metric $m$, and distribution distances use the first-order Wasserstein distance ($W_1$).

At the local scale, we report the average degree ratio (Deg.\ Rat.), the $W_1$ degree distance (Deg.\ W1), and the assortativity difference (Assort.\ Diff.). The average degree ratio is used instead of the edge ratio because some models generate a different number of nodes. The $W_1$ degree distance compares the degree distributions of the generated and reference graphs. Degree assortativity is the correlation between the degrees at the two ends of an edge. It is positive when high-degree nodes connect to each other and negative when they connect to low-degree nodes.

At the higher-order scale, we report the triangle count ratio (Tri.\ Rat.) and the ratio of global clustering coefficients, i.e., transitivity (Clus.\ C.\ Rat.).

At the mesoscopic scale, we report the modularity difference (Mod.\ Diff.) and the ratio of inter-community to intra-community edge counts (Inter/Intra), $\frac{\Gen(\texttt{Inter}/\texttt{Intra})}{G(\texttt{Inter}/\texttt{Intra})}$. Both are computed from a Leiden partition of each graph, independently obtained for the reference and for each sample. Modularity measures how strongly the graph separates into communities, and the Inter/Intra ratio measures how its edges split between and within communities.

At the global scale, we report the characteristic path length ratio (CPL Rat.), where the characteristic path length is the mean shortest-path distance over all reachable node pairs.

A good generator does not need a perfect score on every metric. Rewiring a few edges of the reference scores nearly perfectly on all of them, and it is not a generative model. Some models are also built toward particular metrics: a model conditioned on the reference degree sequence will score well on degree-based measures by construction. We therefore read the metrics as a profile of the generated graph, not as a single objective to maximize.

Structural fidelity results are given in Tables~\ref{tab:citeseer}, \ref{tab:coraml}, \ref{tab:amazonphoto}, \ref{tab:amazoncomputers}, and \ref{tab:flickr}, for the datasets Citeseer, Cora ML, Amazon Photo, Amazon Computers, and Flickr, respectively.

Among the classical models, the mesoscopic metrics separate DC-SBM from the rest. ER, CL, and Config reach Inter/Intra ratios between $2.4$ and $20.5$, and BA exceeds $2.4$ on every graph except Citeseer. CL and Config match the degree distribution and still lose the community structure. DC-SBM is the only classical model that receives a community partition as input. A degree-based model, therefore, does not recover the mesoscopic scale on its own.

DC-SBM is the strongest classical baseline. Its Inter/Intra ratio ranges from $0.80$ to $1.13$ across all five datasets, and it has the lowest modularity difference on three of them. This is expected, since the model is given the Leiden partition of the reference directly and reproduces the community structure it was handed. \method{} starts from the same Leiden partition, so the comparison isolates the edge model: \method{} reaches a comparable Inter/Intra ratio on the four evaluation graphs with a learned edge model and also generates attributes.

Among the models that generate attributes, GenCAT has the lowest degree distance across all datasets, and it completes on all five. On Flickr, its degree distance is $0.020$, and its assortativity difference is $0.002$. Its Inter/Intra ratio is between $1.7$ and $3.0$ on four of the five datasets, so the edges it places correctly at the level of individual nodes are misdistributed between communities. GenCAT takes a class mixing matrix, class-wise attribute statistics, and the degree sequence from the reference as input, so part of what it reproduces is what it was given.

The autoencoder baselines are inconsistent across datasets. GAE produces triangle ratios of $11.6$, $6.1$, and $9.0$ across three of the datasets and $579$ on Flickr, and its degree ratio is exactly one by construction. On those four datasets, GAE matches the edge count and places those edges in dense clusters that the reference does not have. VGAE is closer to triangles but reaches Inter/Intra ratios of $9.9$ on Citeseer and $13.0$ on Amazon Photo.

Completion rates fall with scale. All models run on the two citation graphs. One fails on Amazon Photo, five fail on Amazon Computers, and six fail on Flickr, either due to memory exhaustion or to reaching the time limit. The models that fail either produce a full score matrix or adopt a diffusion process.

The random-walk models trade one scale against another. CELL matches the edge count exactly, and NetGAN is at $0.85$ of it on every graph where it runs. NetGAN's triangle ratio on Citeseer ($1.024$) is among the closest to one, but the Inter/Intra ratios of the two models range from $1.04$ to $17.5$. NetGAN runs out of memory on Amazon Computers and CELL on Flickr.

The diffusion models follow two designs. GraphMaker runs a diffusion process over the whole graph and is inconsistent across its variants: the synchronous one is the better of the two on Citeseer, with a degree ratio of $0.864$ versus $4.284$, and the worse on Amazon Photo, at $5.178$ versus $0.921$. Neither variant completes on Amazon Computers or Flickr. SaGess and LGSG instead diffuse over sampled subgraphs and reassemble them. SaGess is among the closest to the reference in most columns where it runs, and the memorization results show that it does so by reproducing most of the reference edge set. LGSG reassembles by merging nodes with similar embeddings. Its degree ratios of $0.117$ on Amazon Photo and $0.536$ on Cora ML indicate that the merge discards a large part of the edge set, while on Citeseer it overshoots at $1.113$.

SynGen differs from the other baselines in how it sizes the output. It matches the reference edge count but rounds up the node count to the next power of 2, so the same edges are spread over more nodes, and the degree ratio falls between $0.68$ and $0.93$. Its assortativity difference ranges from $0.08$ to $0.25$, among the highest of the models that complete on all five datasets.

On the two citation graphs and Amazon Photo, \method{}'s degree ratio stays near one and its Inter/Intra ratio near the reference. Its assortativity difference on Cora ML is $0.002$, the lowest in that table. Amazon Computers is its weakest result, with a degree ratio of $1.277$, likely the per-node budget bounds the edges a node selects but not the edges it receives, so high-degree nodes gain edges. Across the four datasets, the triangle ratio ranges from $0.28$ to $0.55$, which is consistent with scoring node pairs independently (Section~\ref{sec:structural}). The characteristic path length is between $0.83$ and $0.98$ of the reference, so the generated graphs are slightly more compact. Flickr is an exception to both and is discussed below.

On Flickr, \method{} uses a candidate pool of 500 nodes per cluster pair. Its degree distance of $2.287$ is below that of every learned baseline that completes there. The Inter/Intra ratio, however, falls to $0.097$. The pool covers only a small fraction of the nodes with inter-community edges in the reference, and an edge whose endpoint is outside the pool cannot be generated. Enlarging the pool raises the Inter/Intra ratio but distorts other metrics. At $4000$ nodes, it reaches $0.414$, while the degree distance rises to $10.772$ and the assortativity difference to $0.181$, because the additional edges concentrate on the nodes already in the pool. We therefore report the setting that preserves the degree distribution.

Two measurements reverse on Flickr. The triangle ratio is $4.579$, and the characteristic path length ratio exceeds 1, whereas both are below 1 for the other four datasets. A candidate pool of $2000$ nodes increases the triangle ratio to $122.9$. The reference has $63{,}941$ triangles over $89{,}250$ nodes, against about $1.5$ million over $13{,}752$ nodes for Amazon Computers, so a small absolute excess moves the ratio by a large factor. GAE shows the same sensitivity here, at $579$, compared to $6$ to $12$ on Citeseer, Cora ML, and Amazon Computers.

\begin{table}[t]
\centering
\caption{Amazon Photo structural fidelity results.}
\label{tab:amazonphoto}
\resizebox{\columnwidth}{!}{
\setlength{\tabcolsep}{2pt}
\begin{tabular}{l|c|c|c|c|c|c|c|c}
\toprule
Model  & Deg. Rat. & Deg. W1 & Tri. Rat. & Clus. C. Rat. & Assort. Diff. & Mod. Diff. & Inter/Intra & CPL Rat.\\
\midrule

ER $G(n,p)$ & $0.999 {\scriptscriptstyle \pm 0.003}$ & $18.734 {\scriptscriptstyle \pm 0.055}$ & $0.007 {\scriptscriptstyle \pm 0.000}$ & $0.023 {\scriptscriptstyle \pm 0.000}$ & $0.043 {\scriptscriptstyle \pm 0.003}$ & $0.576 {\scriptscriptstyle \pm 0.002}$ & $14.867 {\scriptscriptstyle \pm 4.432}$ & $0.716 {\scriptscriptstyle \pm 0.000}$ \\
ER $G(n,m)$  & $1.000 {\scriptscriptstyle \pm 0.000}$ & $18.750 {\scriptscriptstyle \pm 0.029}$ & $0.007 {\scriptscriptstyle \pm 0.000}$ & $0.023 {\scriptscriptstyle \pm 0.000}$ & $0.045 {\scriptscriptstyle \pm 0.004}$ & $0.576 {\scriptscriptstyle \pm 0.001}$ & $15.704 {\scriptscriptstyle \pm 3.271}$ & $0.716 {\scriptscriptstyle \pm 0.000}$ \\
WS & $0.964 {\scriptscriptstyle \pm 0.000}$ & $21.240 {\scriptscriptstyle \pm 0.014}$ & $0.817 {\scriptscriptstyle \pm 0.002}$ & $2.968 {\scriptscriptstyle \pm 0.008}$ & $0.041 {\scriptscriptstyle \pm 0.003}$ & $0.104 {\scriptscriptstyle \pm 0.001}$ & $1.153 {\scriptscriptstyle \pm 0.008}$ & $0.899 {\scriptscriptstyle \pm 0.001}$ \\
BA & $1.026 {\scriptscriptstyle \pm 0.000}$ & $7.820 {\scriptscriptstyle \pm 0.020}$ & $0.067 {\scriptscriptstyle \pm 0.001}$ & $0.099 {\scriptscriptstyle \pm 0.001}$ & $0.034 {\scriptscriptstyle \pm 0.003}$ & $0.586 {\scriptscriptstyle \pm 0.001}$ & $19.438 {\scriptscriptstyle \pm 1.753}$ & $0.688 {\scriptscriptstyle \pm 0.001}$ \\
CL & $0.997 {\scriptscriptstyle \pm 0.003}$ & $0.593 {\scriptscriptstyle \pm 0.052}$ & $0.191 {\scriptscriptstyle \pm 0.003}$ & $0.195 {\scriptscriptstyle \pm 0.002}$ & $0.018 {\scriptscriptstyle \pm 0.001}$ & $0.597 {\scriptscriptstyle \pm 0.001}$ & $20.538 {\scriptscriptstyle \pm 2.191}$ & $0.698 {\scriptscriptstyle \pm 0.001}$ \\
Config & $0.981 {\scriptscriptstyle \pm 0.000}$ & $0.582 {\scriptscriptstyle \pm 0.015}$ & $0.137 {\scriptscriptstyle \pm 0.001}$ & $0.157 {\scriptscriptstyle \pm 0.001}$ & $0.010 {\scriptscriptstyle \pm 0.001}$ & $0.594 {\scriptscriptstyle \pm 0.002}$ & $20.407 {\scriptscriptstyle \pm 1.599}$ & $0.704 {\scriptscriptstyle \pm 0.000}$ \\
DC-SBM & $0.910 {\scriptscriptstyle \pm 0.003}$ & $2.797 {\scriptscriptstyle \pm 0.082}$ & $0.375 {\scriptscriptstyle \pm 0.003}$ & $0.585 {\scriptscriptstyle \pm 0.003}$ & $0.103 {\scriptscriptstyle \pm 0.002}$ & $0.005 {\scriptscriptstyle \pm 0.001}$ & $1.128 {\scriptscriptstyle \pm 0.009}$ & $0.911 {\scriptscriptstyle \pm 0.005}$ \\
GenCAT & $0.960 {\scriptscriptstyle \pm 0.001}$ & $1.233 {\scriptscriptstyle \pm 0.046}$ & $0.299 {\scriptscriptstyle \pm 0.008}$ & $0.365 {\scriptscriptstyle \pm 0.009}$ & $0.039 {\scriptscriptstyle \pm 0.003}$ & $0.100 {\scriptscriptstyle \pm 0.006}$ & $1.692 {\scriptscriptstyle \pm 0.261}$ & $0.779 {\scriptscriptstyle \pm 0.002}$ \\
GAE & $1.000 {\scriptscriptstyle \pm 0.000}$ & $5.150 {\scriptscriptstyle \pm 0.044}$ & $0.811 {\scriptscriptstyle \pm 0.005}$ & $0.911 {\scriptscriptstyle \pm 0.003}$ & $0.112 {\scriptscriptstyle \pm 0.001}$ & $0.258 {\scriptscriptstyle \pm 0.004}$ & $3.752 {\scriptscriptstyle \pm 0.246}$ & $0.750 {\scriptscriptstyle \pm 0.001}$ \\
VGAE & $1.000 {\scriptscriptstyle \pm 0.000}$ & $5.006 {\scriptscriptstyle \pm 0.190}$ & $0.249 {\scriptscriptstyle \pm 0.003}$ & $0.436 {\scriptscriptstyle \pm 0.003}$ & $0.085 {\scriptscriptstyle \pm 0.004}$ & $0.482 {\scriptscriptstyle \pm 0.003}$ & $13.042 {\scriptscriptstyle \pm 1.025}$ & $0.740 {\scriptscriptstyle \pm 0.001}$ \\
NetGAN & $0.850 {\scriptscriptstyle \pm 0.000}$ & $26.537 {\scriptscriptstyle \pm 0.033}$ & $0.877 {\scriptscriptstyle \pm 0.003}$ & $0.042 {\scriptscriptstyle \pm 0.000}$ & $0.680 {\scriptscriptstyle \pm 0.002}$ & $0.615 {\scriptscriptstyle \pm 0.002}$ & $17.509 {\scriptscriptstyle \pm 1.158}$ & $0.561 {\scriptscriptstyle \pm 0.001}$ \\
CELL & $1.000 {\scriptscriptstyle \pm 0.000}$ & $2.584 {\scriptscriptstyle \pm 0.055}$ & $0.626 {\scriptscriptstyle \pm 0.004}$ & $0.840 {\scriptscriptstyle \pm 0.005}$ & $0.015 {\scriptscriptstyle \pm 0.002}$ & $0.009 {\scriptscriptstyle \pm 0.001}$ & $1.105 {\scriptscriptstyle \pm 0.020}$ & $0.842 {\scriptscriptstyle \pm 0.002}$ \\
GraphMaker S & $5.178 {\scriptscriptstyle \pm 0.169}$ & $130.322 {\scriptscriptstyle \pm 5.262}$ & $25.411 {\scriptscriptstyle \pm 2.984}$ & $1.519 {\scriptscriptstyle \pm 0.055}$ & $0.636 {\scriptscriptstyle \pm 0.018}$ & $0.158 {\scriptscriptstyle \pm 0.030}$ & $0.274 {\scriptscriptstyle \pm 0.011}$ & $0.661 {\scriptscriptstyle \pm 0.002}$ \\
GraphMaker A & $0.921 {\scriptscriptstyle \pm 0.038}$ & $4.961 {\scriptscriptstyle \pm 0.448}$ & $0.845 {\scriptscriptstyle \pm 0.129}$ & $1.230 {\scriptscriptstyle \pm 0.076}$ & $0.160 {\scriptscriptstyle \pm 0.032}$ & $0.009 {\scriptscriptstyle \pm 0.006}$ & $0.629 {\scriptscriptstyle \pm 0.039}$ & $0.936 {\scriptscriptstyle \pm 0.006}$ \\
SaGess & \multicolumn{4}{c}{TO} & \multicolumn{4}{c}{TO} \\
LGSG & $0.117 {\scriptscriptstyle \pm 0.001}$ & $28.963 {\scriptscriptstyle \pm 0.044}$ & $0.010 {\scriptscriptstyle \pm 0.000}$ & $0.005 {\scriptscriptstyle \pm 0.000}$ & $0.288 {\scriptscriptstyle \pm 0.006}$ & $0.216 {\scriptscriptstyle \pm 0.006}$ & $3.970 {\scriptscriptstyle \pm 0.086}$ & $0.541 {\scriptscriptstyle \pm 0.002}$ \\
SynGen & $0.934 {\scriptscriptstyle \pm 0.000}$ & $7.069 {\scriptscriptstyle \pm 1.869}$ & $0.088 {\scriptscriptstyle \pm 0.050}$ & $0.181 {\scriptscriptstyle \pm 0.066}$ & $0.137 {\scriptscriptstyle \pm 0.070}$ & $0.335 {\scriptscriptstyle \pm 0.043}$ & $4.997 {\scriptscriptstyle \pm 1.184}$ & $0.762 {\scriptscriptstyle \pm 0.008}$ \\

\midrule
\textbf{\method} & $0.966 {\scriptscriptstyle \pm 0.002}$ & $5.417 {\scriptscriptstyle \pm 0.069}$ & $0.294 {\scriptscriptstyle \pm 0.003}$ & $0.273 {\scriptscriptstyle \pm 0.003}$ & $0.024 {\scriptscriptstyle \pm 0.001}$ & $0.043 {\scriptscriptstyle \pm 0.003}$ & $1.159 {\scriptscriptstyle \pm 0.026}$ & $0.829 {\scriptscriptstyle \pm 0.003}$ \\

\bottomrule
\end{tabular}
}
\end{table}

\begin{table}[t]
\centering
\caption{Amazon Computers structural fidelity results.}
\label{tab:amazoncomputers}
\resizebox{\columnwidth}{!}{
\setlength{\tabcolsep}{2pt}
\begin{tabular}{l|c|c|c|c|c|c|c|c}
\toprule
Model  & Deg. Rat. & Deg. W1 & Tri. Rat. & Clus. C. Rat. & Assort. Diff. & Mod. Diff. & Inter/Intra & CPL Rat.\\
\midrule
ER $G(n,p)$ & $1.000 {\scriptscriptstyle \pm 0.002}$ & $24.577 {\scriptscriptstyle \pm 0.033}$ & $0.005 {\scriptscriptstyle \pm 0.000}$ & $0.024 {\scriptscriptstyle \pm 0.000}$ & $0.056 {\scriptscriptstyle \pm 0.002}$ & $0.487 {\scriptscriptstyle \pm 0.002}$ & $11.048 {\scriptscriptstyle \pm 1.701}$ & $0.872 {\scriptscriptstyle \pm 0.000}$ \\
ER $G(n,m)$ & $1.000 {\scriptscriptstyle \pm 0.000}$ & $24.562 {\scriptscriptstyle \pm 0.022}$ & $0.005 {\scriptscriptstyle \pm 0.000}$ & $0.024 {\scriptscriptstyle \pm 0.000}$ & $0.056 {\scriptscriptstyle \pm 0.003}$ & $0.487 {\scriptscriptstyle \pm 0.001}$ & $10.704 {\scriptscriptstyle \pm 1.886}$ & $0.872 {\scriptscriptstyle \pm 0.000}$ \\
WS & $1.007 {\scriptscriptstyle \pm 0.000}$ & $27.549 {\scriptscriptstyle \pm 0.013}$ & $1.005 {\scriptscriptstyle \pm 0.002}$ & $4.918 {\scriptscriptstyle \pm 0.012}$ & $0.053 {\scriptscriptstyle \pm 0.001}$ & $0.215 {\scriptscriptstyle \pm 0.001}$ & $0.611 {\scriptscriptstyle \pm 0.008}$ & $1.081 {\scriptscriptstyle \pm 0.001}$ \\
BA & $1.005 {\scriptscriptstyle \pm 0.000}$ & $11.047 {\scriptscriptstyle \pm 0.020}$ & $0.052 {\scriptscriptstyle \pm 0.001}$ & $0.113 {\scriptscriptstyle \pm 0.001}$ & $0.049 {\scriptscriptstyle \pm 0.001}$ & $0.494 {\scriptscriptstyle \pm 0.001}$ & $10.036 {\scriptscriptstyle \pm 1.078}$ & $0.839 {\scriptscriptstyle \pm 0.000}$ \\
CL & $0.996 {\scriptscriptstyle \pm 0.002}$ & $0.606 {\scriptscriptstyle \pm 0.038}$ & $0.287 {\scriptscriptstyle \pm 0.002}$ & $0.307 {\scriptscriptstyle \pm 0.002}$ & $0.016 {\scriptscriptstyle \pm 0.001}$ & $0.508 {\scriptscriptstyle \pm 0.001}$ & $11.379 {\scriptscriptstyle \pm 0.784}$ & $0.836 {\scriptscriptstyle \pm 0.001}$ \\
Config & $0.978 {\scriptscriptstyle \pm 0.000}$ & $0.797 {\scriptscriptstyle \pm 0.006}$ & $0.199 {\scriptscriptstyle \pm 0.001}$ & $0.246 {\scriptscriptstyle \pm 0.001}$ & $0.013 {\scriptscriptstyle \pm 0.000}$ & $0.506 {\scriptscriptstyle \pm 0.001}$ & $9.857 {\scriptscriptstyle \pm 0.865}$ & $0.843 {\scriptscriptstyle \pm 0.001}$ \\
DC-SBM & $0.922 {\scriptscriptstyle \pm 0.003}$ & $2.797 {\scriptscriptstyle \pm 0.090}$ & $0.388 {\scriptscriptstyle \pm 0.002}$ & $0.668 {\scriptscriptstyle \pm 0.003}$ & $0.075 {\scriptscriptstyle \pm 0.001}$ & $0.004 {\scriptscriptstyle \pm 0.001}$ & $0.796 {\scriptscriptstyle \pm 0.020}$ & $0.966 {\scriptscriptstyle \pm 0.001}$ \\

GenCAT & $0.956 {\scriptscriptstyle \pm 0.001}$ & $1.567 {\scriptscriptstyle \pm 0.044}$ & $0.330 {\scriptscriptstyle \pm 0.009}$ & $0.480 {\scriptscriptstyle \pm 0.012}$ & $0.044 {\scriptscriptstyle \pm 0.003}$ & $0.074 {\scriptscriptstyle \pm 0.004}$ & $0.924 {\scriptscriptstyle \pm 0.102}$ & $0.917 {\scriptscriptstyle \pm 0.002}$ \\

GAE & $1.000 {\scriptscriptstyle \pm 0.000}$ & $30.242 {\scriptscriptstyle \pm 0.014}$ & $9.043 {\scriptscriptstyle \pm 0.008}$ & $1.936 {\scriptscriptstyle \pm 0.003}$ & $0.439 {\scriptscriptstyle \pm 0.000}$ & $0.233 {\scriptscriptstyle \pm 0.000}$ & $1.398 {\scriptscriptstyle \pm 0.004}$ & $0.665 {\scriptscriptstyle \pm 0.000}$ \\
VGAE & $1.000 {\scriptscriptstyle \pm 0.000}$ & $6.971 {\scriptscriptstyle \pm 0.116}$ & $0.275 {\scriptscriptstyle \pm 0.001}$ & $0.639 {\scriptscriptstyle \pm 0.002}$ & $0.083 {\scriptscriptstyle \pm 0.003}$ & $0.389 {\scriptscriptstyle \pm 0.004}$ & $7.505 {\scriptscriptstyle \pm 0.528}$ & $0.902 {\scriptscriptstyle \pm 0.001}$ \\

NetGAN & \multicolumn{4}{c}{OOM} & \multicolumn{4}{c}{OOM} \\

CELL & $1.000 {\scriptscriptstyle \pm 0.000}$ & $1.299 {\scriptscriptstyle \pm 0.023}$ & $0.649 {\scriptscriptstyle \pm 0.003}$ & $0.812 {\scriptscriptstyle \pm 0.003}$ & $0.008 {\scriptscriptstyle \pm 0.001}$ & $0.025 {\scriptscriptstyle \pm 0.001}$ & $1.044 {\scriptscriptstyle \pm 0.066}$ & $0.982 {\scriptscriptstyle \pm 0.001}$ \\

GraphMaker S & \multicolumn{4}{c}{TO} & \multicolumn{4}{c}{TO} \\
GraphMaker A & \multicolumn{4}{c}{TO} & \multicolumn{4}{c}{TO} \\
SaGess & \multicolumn{4}{c}{TO} & \multicolumn{4}{c}{TO} \\
LGSG & \multicolumn{4}{c}{TO} & \multicolumn{4}{c}{TO} \\

SynGen & $0.839 {\scriptscriptstyle \pm 0.000}$ & $9.535 {\scriptscriptstyle \pm 2.205}$ & $0.063 {\scriptscriptstyle \pm 0.027}$ & $0.223 {\scriptscriptstyle \pm 0.078}$ & $0.248 {\scriptscriptstyle \pm 0.092}$ & $0.236 {\scriptscriptstyle \pm 0.034}$ & $1.650 {\scriptscriptstyle \pm 0.471}$ & $0.970 {\scriptscriptstyle \pm 0.015}$ \\
\midrule
\textbf{\method} & $1.277 {\scriptscriptstyle \pm 0.001}$ & $11.084 {\scriptscriptstyle \pm 0.023}$ & $0.549 {\scriptscriptstyle \pm 0.002}$ & $0.498 {\scriptscriptstyle \pm 0.002}$ & $0.092 {\scriptscriptstyle \pm 0.001}$ & $0.073 {\scriptscriptstyle \pm 0.003}$ & $0.945 {\scriptscriptstyle \pm 0.085}$ & $0.978 {\scriptscriptstyle \pm 0.001}$ \\
\bottomrule
\end{tabular}
}
\end{table}

\begin{table}[t]
\centering
\caption{Flickr structural fidelity results.}
\label{tab:flickr}
\resizebox{\columnwidth}{!}{
\setlength{\tabcolsep}{2pt}
\begin{tabular}{l|c|c|c|c|c|c|c|c}
\toprule
Model  & Deg. Rat. & Deg. W1 & Tri. Rat. & Clus. C. Rat. & Assort. Diff. & Mod. Diff. & Inter/Intra & CPL Rat.\\

\midrule
ER $G(n,p)$ & $1.000 {\scriptscriptstyle \pm 0.002}$ & $5.127 {\scriptscriptstyle \pm 0.014}$ & $0.003 {\scriptscriptstyle \pm 0.000}$ & $0.029 {\scriptscriptstyle \pm 0.002}$ & $0.044 {\scriptscriptstyle \pm 0.001}$ & $0.248 {\scriptscriptstyle \pm 0.002}$ & $3.551 {\scriptscriptstyle \pm 0.151}$ & $1.221 {\scriptscriptstyle \pm 0.001}$ \\
ER $G(n,m)$ & $1.000 {\scriptscriptstyle \pm 0.000}$ & $5.127 {\scriptscriptstyle \pm 0.005}$ & $0.003 {\scriptscriptstyle \pm 0.000}$ & $0.029 {\scriptscriptstyle \pm 0.002}$ & $0.044 {\scriptscriptstyle \pm 0.002}$ & $0.249 {\scriptscriptstyle \pm 0.002}$ & $3.549 {\scriptscriptstyle \pm 0.196}$ & $1.221 {\scriptscriptstyle \pm 0.000}$ \\
WS & $0.992 {\scriptscriptstyle \pm 0.000}$ & $6.020 {\scriptscriptstyle \pm 0.002}$ & $10.167 {\scriptscriptstyle \pm 0.019}$ & $122.913 {\scriptscriptstyle \pm 0.231}$ & $0.035 {\scriptscriptstyle \pm 0.002}$ & $0.353 {\scriptscriptstyle \pm 0.000}$ & $0.218 {\scriptscriptstyle \pm 0.001}$ & $1.815 {\scriptscriptstyle \pm 0.003}$ \\
BA & $0.992 {\scriptscriptstyle \pm 0.000}$ & $1.565 {\scriptscriptstyle \pm 0.008}$ & $0.063 {\scriptscriptstyle \pm 0.003}$ & $0.211 {\scriptscriptstyle \pm 0.006}$ & $0.030 {\scriptscriptstyle \pm 0.001}$ & $0.243 {\scriptscriptstyle \pm 0.001}$ & $2.468 {\scriptscriptstyle \pm 0.139}$ & $1.002 {\scriptscriptstyle \pm 0.002}$ \\
CL & $0.999 {\scriptscriptstyle \pm 0.001}$ & $1.004 {\scriptscriptstyle \pm 0.008}$ & $1.271 {\scriptscriptstyle \pm 0.015}$ & $1.313 {\scriptscriptstyle \pm 0.017}$ & $0.026 {\scriptscriptstyle \pm 0.000}$ & $0.252 {\scriptscriptstyle \pm 0.001}$ & $2.473 {\scriptscriptstyle \pm 0.064}$ & $0.897 {\scriptscriptstyle \pm 0.001}$ \\
Config & $0.996 {\scriptscriptstyle \pm 0.000}$ & $0.043 {\scriptscriptstyle \pm 0.001}$ & $0.877 {\scriptscriptstyle \pm 0.009}$ & $0.980 {\scriptscriptstyle \pm 0.011}$ & $0.023 {\scriptscriptstyle \pm 0.000}$ & $0.246 {\scriptscriptstyle \pm 0.000}$ & $2.403 {\scriptscriptstyle \pm 0.032}$ & $0.892 {\scriptscriptstyle \pm 0.000}$ \\
DC-SBM & $0.976 {\scriptscriptstyle \pm 0.001}$ & $1.152 {\scriptscriptstyle \pm 0.011}$ & $1.443 {\scriptscriptstyle \pm 0.019}$ & $2.065 {\scriptscriptstyle \pm 0.034}$ & $0.024 {\scriptscriptstyle \pm 0.000}$ & $0.021 {\scriptscriptstyle \pm 0.008}$ & $0.811 {\scriptscriptstyle \pm 0.087}$ & $0.951 {\scriptscriptstyle \pm 0.001}$ \\

GenCAT & $0.998 {\scriptscriptstyle \pm 0.000}$ & $0.020 {\scriptscriptstyle \pm 0.001}$ & $0.083 {\scriptscriptstyle \pm 0.009}$ & $0.083 {\scriptscriptstyle \pm 0.009}$ & $0.002 {\scriptscriptstyle \pm 0.000}$ & $0.213 {\scriptscriptstyle \pm 0.008}$ & $2.297 {\scriptscriptstyle \pm 0.250}$ & $0.942 {\scriptscriptstyle \pm 0.001}$ \\

GAE & $1.000 {\scriptscriptstyle \pm 0.000}$ & $12.832 {\scriptscriptstyle \pm 0.002}$ & $579.284 {\scriptscriptstyle \pm 0.231}$ & $33.066 {\scriptscriptstyle \pm 0.041}$ & $0.067 {\scriptscriptstyle \pm 0.000}$ & $0.011 {\scriptscriptstyle \pm 0.002}$ & $0.228 {\scriptscriptstyle \pm 0.049}$ & $0.535 {\scriptscriptstyle \pm 0.000}$ \\
VGAE & $1.000 {\scriptscriptstyle \pm 0.000}$ & $6.172 {\scriptscriptstyle \pm 0.017}$ & $4.120 {\scriptscriptstyle \pm 0.034}$ & $8.705 {\scriptscriptstyle \pm 0.054}$ & $0.110 {\scriptscriptstyle \pm 0.006}$ & $0.208 {\scriptscriptstyle \pm 0.003}$ & $1.910 {\scriptscriptstyle \pm 0.109}$ & $0.951 {\scriptscriptstyle \pm 0.001}$ \\

NetGAN & \multicolumn{4}{c}{OOM} & \multicolumn{4}{c}{OOM} \\
CELL & \multicolumn{4}{c}{OOM} & \multicolumn{4}{c}{OOM} \\
GraphMaker S & \multicolumn{4}{c}{OOM} & \multicolumn{4}{c}{OOM} \\
GraphMaker A & \multicolumn{4}{c}{OOM} & \multicolumn{4}{c}{OOM} \\
SaGess & \multicolumn{4}{c}{OOM} & \multicolumn{4}{c}{OOM} \\
LGSG & \multicolumn{4}{c}{TO} & \multicolumn{4}{c}{TO} \\

SynGen & $0.681 {\scriptscriptstyle \pm 0.000}$ & $3.885 {\scriptscriptstyle \pm 0.254}$ & $0.069 {\scriptscriptstyle \pm 0.044}$ & $0.451 {\scriptscriptstyle \pm 0.220}$ & $0.223 {\scriptscriptstyle \pm 0.064}$ & $0.112 {\scriptscriptstyle \pm 0.039}$ & $1.154 {\scriptscriptstyle \pm 0.511}$ & $1.276 {\scriptscriptstyle \pm 0.057}$ \\
\midrule
\textbf{\method} & $0.831 {\scriptscriptstyle \pm 0.000}$ & $2.287 {\scriptscriptstyle \pm 0.005}$ & $4.579 {\scriptscriptstyle \pm 0.114}$ & $3.333 {\scriptscriptstyle \pm 0.084}$ & $0.026 {\scriptscriptstyle \pm 0.000}$ & $0.273 {\scriptscriptstyle \pm 0.001}$ & $0.097 {\scriptscriptstyle \pm 0.001}$ & $1.259 {\scriptscriptstyle \pm 0.003}$ \\

\bottomrule
\end{tabular}
}
\end{table}

\subsection{Memorization Evaluation}
\label{app:memorization}

In the multi-graph setting, a generated graph is memorized if it is isomorphic or nearly isomorphic to a training graph. With a single reference, this definition does not apply: the model is trained to produce a graph that resembles the reference, so some overlap is expected. A model can also score perfectly on every criterion except memorization by copying the reference. 

Even a model that generates every edge independently shares some edges with the reference if it reproduces the reference degree sequence and community structure, so a non-zero overlap alone does not indicate memorization.

We also measure self-overlap, the mean overlap between pairs of samples of the same model. It exposes a second failure mode: low edge overlap with high self-overlap means the model produces nearly the same graph every time.

\subsubsection{Alignment}

Memorization metrics require node correspondence between graphs. Graph structure is invariant under node permutations, so that the same graph can be represented in different node orderings. The overlap between $G(\pi_1)$ and $G(\pi_2)$ therefore depends on the permutations $\pi_1$ and $\pi_2$, although both represent the same graph.

The models fall into three alignment classes. Unaligned models (ER, WS, BA, VGAE, GraphMaker, LGSG, and SynGen) assign node identities arbitrarily, so their overlaps carry no meaning, and we exclude them. Degree-aligned models (CL, Config, and GenCAT) are aligned by matching nodes of equal degree. This alignment comes from the degree sequence, not from the model, so it gives a lower bound on their overlap. Node-aligned models (DC-SBM, GAE, CELL, NetGAN, SaGess, and \method{}) keep the reference node identities, and all measures are defined directly.

\subsubsection{Memorization Metrics}

Let $E$ be the undirected edge set of the reference, $\hat{E}^{(i)}$ that of the $i$-th generated sample, and $S$ the number of samples ($S=10$ throughout). All sets exclude self-loops and duplicate edges.
 
\textbf{Edge overlap} is the Jaccard similarity
\begin{equation*}
\mathrm{EO} = \frac{1}{S}\sum_{i=1}^{S} \frac{|E \cap \hat{E}^{(i)}|}{|E \cup \hat{E}^{(i)}|} .
\end{equation*}
 
\textbf{Self-overlap} is the same similarity taken over sample pairs,
\begin{equation*}
\mathrm{SO} = \binom{S}{2}^{-1} \sum_{i<j} \frac{|\hat{E}^{(i)} \cap \hat{E}^{(j)}|}{|\hat{E}^{(i)} \cup \hat{E}^{(j)}|} .
\end{equation*}
 
Neighborhood memorization is computed per node. Let $\mathcal{N}(v)$ and $\hat{\mathcal{N}}(v)$ be the neighbor sets of $v$ in the reference and in a sample, and
\begin{equation*}
J(v) = \frac{|\mathcal{N}(v) \cap \hat{\mathcal{N}}(v)|}{|\mathcal{N}(v) \cup \hat{\mathcal{N}}(v)|} .
\end{equation*}
We report the mean of $J(v)$ over nodes and samples (NM) and the fraction of nodes with $J(v) \geq 0.9$ (Frac.). Nodes isolated in both the reference and the sample are excluded, since $J(v)$ is undefined for them.
 
\textbf{Attribute-level memorization.} The four metrics above measure structural memorization. For attributes, we compute for each generated node the Euclidean distances $d_1$ and $d_2$ to its nearest and second-nearest reference nodes. We report the median of $d_1/d_2$ as the nearest-neighbor distance ratio (NNDR), and the fraction of generated nodes whose $d_1$ is below a threshold (Leak), over 10 samples per model. A ratio near zero means the node sits on one reference node and not between several, which indicates copying at the attribute level. The threshold is set to the fifth percentile of the reference nearest-neighbor distances. Hence, a model whose generated attributes follow the same distribution as the reference yields a fraction near $0.05$ by construction. Since the threshold is derived from the reference, the fraction does not depend on the scale of the attribute space. It does depend on the generated attributes following the same marginal distribution. When they do not, the fraction collapses toward zero, or toward one when the generated attributes fall inside a dense region of the reference, and the measure no longer says anything about copying. We therefore give two marginal statistics alongside it, the mean number of non-zero attributes per node and the Spearman correlation between the per-attribute frequencies of the generated and reference graphs. The measurement requires no node identity, as it compares two sets of attribute points.

\subsubsection{Results}

Structural memorization results are in Table~\ref{app:structuralmemorization} and attribute memorization results in Table~\ref{app:featurememorization}.

\begin{table}[t]
\centering
\caption{Structural memorization results. $^\dagger$NetGAN is trained and evaluated on the largest connected component, so its values are computed over a smaller node set than the other rows. EO is edge overlap, SO is self-overlap, and NM is neighborhood memorization. Frac.\ is the fraction of nodes with $J(v) \geq 0.9$.}
\label{app:structuralmemorization}
\resizebox{\columnwidth}{!}{
\setlength{\tabcolsep}{2pt}
\begin{tabular}{l|c|cccc|cccc|cccc}
\toprule
\multirow{2}{*}{Model}  & \multirow{2}{*}{Alignment} & \multicolumn{4}{c}{Citeseer} & \multicolumn{4}{c}{Cora ML} & \multicolumn{4}{c}{Amazon Photo}\\
\cmidrule(lr){3-6} 
\cmidrule(lr){7-10} 
\cmidrule(lr){11-14} 
& & EO & SO & NM & Frac. & EO & SO & NM & Frac. & EO & SO & NM & Frac. \\
\midrule

CL & Degree & $0.003$ & $0.003$ & $0.001$ & $0.000$ & $0.007$ & $0.009$ & $0.003$ & $0.000$ & $0.019$ & $0.020$ & $0.007$ & $0.000$ \\
Config & Degree & $0.003$ & $0.002$ & $0.001$ & $0.000$ & $0.007$ & $0.008$ & $0.003$ & $0.000$ & $0.017$ & $0.016$ & $0.007$ & $0.000$ \\
GenCAT & Degree & $0.002$ & $0.004$ & $0.001$ & $0.000$ & $0.005$ & $0.014$ & $0.002$ & $0.000$ & $0.015$ & $0.028$ & $0.007$ & $0.000$ \\
\midrule
DC-SBM & Node &$0.084$ & $0.063$ & $0.140$ & $0.085$ & $0.052$ & $0.046$ & $0.050$ & $0.015$ & $0.075$ & $0.066$ & $0.046$ & $0.002$ \\
GAE & Node & $0.227$ & $0.750$ & $0.125$ & $0.017$ & $0.191$ & $0.749$ & $0.106$ & $0.004$ & $0.174$ & $0.743$ & $0.075$ & $0.000$ \\
CELL & Node & $0.384$ & $0.258$ & $0.500$ & $0.198$ & $0.442$ & $0.284$ & $0.516$ & $0.129$ & $0.337$ & $0.237$ & $0.340$ & $0.007$ \\
NetGAN$^\dagger$ & Node & $0.228$ & $0.142$ & $0.208$ & $0.034$ & $0.174$ & $0.094$ & $0.151$ & $0.015$ & $0.027$ & $0.801$ & $0.025$ & $0.000$ \\
SaGess & Node & $0.910$ & $0.847$ & $0.904$ & $0.818$ & $0.815$ & $0.735$ & $0.795$ & $0.524$ & \multicolumn{4}{c}{TO} \\
\midrule
\textbf{\method} & Node & $0.102$ & $0.108$ & $0.254$ & $0.183$ & $0.038$ & $0.066$ & $0.068$ & $0.034$ & $0.040$ & $0.068$ & $0.035$ & $0.005$ \\

\midrule
\midrule
\multirow{2}{*}{Model}  & \multirow{2}{*}{Alignment} & \multicolumn{4}{c}{Amazon Computers} & \multicolumn{4}{c}{Flickr} \\
\cmidrule(lr){3-6} 
\cmidrule(lr){7-10} 

& & EO & SO & NM & Frac. & EO & SO & NM & Frac. \\
\midrule
CL & Degree & $0.021$ & $0.022$ & $0.007$ & $0.000$ & $0.001$ & $0.004$ & $0.001$ & $0.000$ \\
Config & Degree & $0.019$ & $0.017$ & $0.006$ & $0.000$ & $0.001$ & $0.003$ & $0.001$ & $0.000$ \\
GenCAT & Degree & $0.015$ & $0.029$ & $0.005$ & $0.000$ & $0.001$ & $0.001$ & $0.001$ & $0.000$ \\
\midrule
DC-SBM & Node & $0.056$ & $0.051$ & $0.031$ & $0.002$ & $0.007$ & $0.009$ & $0.005$ & $0.000$ \\
GAE & Node & $0.102$ & $0.820$ & $0.030$ & $0.000$ & $0.008$ & $0.951$ & $0.002$ & $0.000$ \\
CELL & Node & $0.348$ & $0.232$ & $0.393$ & $0.018$ & \multicolumn{4}{c}{OOM} \\
NetGAN$^\dagger$ & Node & \multicolumn{4}{c}{OOM} & \multicolumn{4}{c}{OOM} \\
SaGess & Node & \multicolumn{4}{c}{TO} & \multicolumn{4}{c}{OOM} \\
\midrule
\textbf{\method} & Node & $0.032$ & $0.111$ & $0.024$ & $0.004$ & $0.003$ & $0.121$ & $0.004$ & $0.000$ \\

\bottomrule
\end{tabular}
}
\end{table}

\paragraph{Structural memorization.}

The alignment classes separate clearly. The degree-aligned models score below $0.03$ on every measure and dataset, and none has a memorized node. This is expected from models that reproduce a degree sequence and nothing else, and the same models distort the community structure in the fidelity experiments. Low overlap alone is therefore not evidence of a good generator.

Self-overlap exposes two models. GAE has an edge overlap between $0.008$ and $0.227$ and a self-overlap between $0.743$ and $0.951$ across the five datasets, indicating that its samples are close copies of each other. This is expected, since GAE has no prior and generates by reconstructing the reference. NetGAN shows the same pattern on Amazon Photo, with an overlap of $0.027$ against a self-overlap of $0.801$, though not on the citation graphs. A measure of overlap with the reference alone would have scored both of these as generalizing.

SaGess memorizes directly. Its edge overlap is $0.910$ on Citeseer and $0.815$ on Cora ML, and $0.818$ and $0.524$ of its nodes keep their reference neighborhood almost exactly. It reconstructs the reference, which explains its structural fidelity. CELL sits between the two extremes, with an edge overlap between $0.337$ and $0.442$ and a mean neighborhood Jaccard of about $0.5$ on the citation graphs, an amount set by its stopping rule.

\method{} has an edge overlap between $0.003$ and $0.102$, and its self-overlap exceeds its edge overlap on all five datasets. The reference is therefore no closer to a generated graph than two generated graphs are to each other, and the gap generally widens with graph size: on Flickr, the edge overlap is $0.003$ against a self-overlap of $0.121$. Its memorized-node fraction is at most $0.034$ on four datasets. Citeseer is the exception at $0.183$, and the analysis below indicates that this largely follows from its component structure.

\paragraph{Attribute memorization.}

NNDR does not discriminate between models. Every model on every dataset returns a value between $0.969$ and $1.000$, including the one with a leak fraction of $0.273$. In a sparse high-dimensional binary attribute space, a generated node shares almost no attributes with any reference node, so the Euclidean distance is set by the number of non-zero attributes rather than by the overlap, and the nearest and second-nearest reference nodes sit at nearly the same distance. Therefore, the ratio carries no information here. We report it for completeness.

The leak fraction is interpretable only when the generated attribute marginals match the reference, and two models fall outside that condition. GenCAT generates continuous attributes with a frequency correlation below $0.3$, so its zero leak reflects the mismatch. On the citation graphs, SynGen generates about 50 times (Citeseer) and 30 times (Cora ML) as many non-zero attributes per node as the reference, with frequency correlations of $0.465$ and $0.389$, respectively. On the Amazon graphs, the marginals are closer (about twice the density, correlation above $0.93$). GenCAT's zero leak on every dataset and SynGen's zero leak on the citation graphs therefore carry no information about copying.

Among the models whose marginals match the reference, the asynchronous variant of GraphMaker has the highest leak, at $0.273$ on Citeseer and $0.074$ on Cora ML, against $0.012$ and $0.000$ for the synchronous variant. Its feature-only accuracy is also above the reference (Table~\ref{tab:homophily_properties}), which is consistent with attributes close to individual reference nodes.

\method{} reproduces the attribute marginals closely on every dataset. Its non-zero attribute count is within a tenth of the reference on the four binary datasets, and its attribute-frequency correlation ranges from $0.858$ to $0.981$, the highest of all models on Cora ML and Flickr. The leak remains at or below $0.002$ everywhere except Flickr, so the attributes it generates are close to the reference distribution but not to any individual reference node.

On Flickr, the attributes are continuous, and the leak fraction inverts: \method{} and GenCAT both reach $1.000$, while SynGen stays at $0.000$. The reference has a mean L1 norm of $431.0$ per node, against $328.7$ for \method{} and $250.0$ for GenCAT, so both generate smaller vectors. Vectors with a smaller norm lie closer to the center of the reference point cloud. A point there is closer to its nearest reference node than most reference nodes are to their own nearest neighbor, so every generated node falls below the threshold. SynGen generates larger vectors ($656.0$) and has zero leak. This is the marginal mismatch described above in the other direction: shrunken attributes cross the threshold everywhere, and inflated ones never do. The NNDR of \method{} on Flickr is $0.981$, far from the value near zero that copying produces, which supports this reading. We mark the Flickr leak values as uninterpretable.

\begin{table}[t]
\centering
\caption{Attribute memorization. The first row of each block is the reference graph. Feat. Den. is the mean number of non-zero attributes per node for binary attributes and the mean L1 norm per node for continuous attributes; it is left empty where the generated attribute type differs from the reference, as the two quantities are then on different scales. Feat. $\rho$ is the Spearman correlation between the column means of the generated and the reference attribute matrix. For binary attributes, the column mean is the frequency of the attribute. The reference row carries no Feat. $\rho$, NNDR, or Leak value, as all three are defined between a generated graph and the reference. $^\ast$ Leak value not interpretable because the generated attribute marginals do not match the reference.}
\label{app:featurememorization}
\resizebox{\columnwidth}{!}{
\setlength{\tabcolsep}{3pt}
\begin{tabular}{l r c c c c | r c c c c}
\toprule
Model & Feat. Type & Feat. Den. & Feat. $\rho$ & NNDR & Leak & Feat. Type & Feat. Den. & Feat. $\rho$ & NNDR & Leak \\
\midrule
& \multicolumn{5}{c}{Citeseer} & \multicolumn{5}{c}{Cora ML} \\
\midrule
\textbf{Reference} & binary & 31.6 & -- & -- & -- & binary & 50.5 & -- & -- & -- \\
GenCAT & cont. & -- & 0.298 & $1.000$ & $0.000^\ast$ & cont. & -- & 0.082 & $0.999$ & $0.000^\ast$ \\
GraphMaker S & binary & 23.1 & 0.873 & $1.000$ & $0.012$ & binary & 33.8 & 0.922 & $0.988$ & $0.000$ \\
GraphMaker A & binary & 21.1 & 0.860  & $1.000$ & $0.273$ & binary & 26.1 & 0.898 & $0.969$ & $0.074$ \\
SynGen & binary & 1561.9 & 0.465 & $1.000$ & $0.000^\ast$ & binary & 1584.7 & 0.389 & $0.999$ & $0.000^\ast$ \\
\method & binary & 28.8 & 0.858 & $1.000$ & $0.000$ & binary & 53.2 & 0.949 & $0.992$ & $0.000$ \\

\midrule
& \multicolumn{5}{c}{Amazon Photo} & \multicolumn{5}{c}{Amazon Computers} \\
\midrule
\textbf{Reference} & binary & 258.8 & -- & -- & -- & binary & 267.2 & -- & -- & -- \\
GenCAT & cont. & -- & 0.013 & $0.999$ & $0.000^\ast$ & cont. & -- & 0.006 & $0.999$ & $0.000^\ast$ \\
GraphMaker S & binary & 263.0 & 0.979 & $0.996$ & $0.000$ & \multicolumn{5}{c}{TO} \\
GraphMaker A & binary & 235.4 & 0.989 & $0.990$ & $0.004$ & \multicolumn{5}{c}{TO} \\
SynGen & binary & 496.0 & 0.935 & $0.998$ & $0.000$ & binary & 510.9 & 0.964 & $0.998$ & $0.000$ \\
\method & binary & 270.5 & 0.963 & $0.996$ & $0.002$ & binary & 264.4 & 0.955 & $0.997$ & $0.002$ \\

\midrule
& \multicolumn{5}{c}{Flickr} \\
\midrule
\textbf{Reference} & cont. & 431.0 & -- & -- & -- \\
GenCAT & cont. & 250.0 & 0.002 & $0.999$ & $1.000^\ast$ \\
GraphMaker S & \multicolumn{5}{c}{OOM} \\
GraphMaker A & \multicolumn{5}{c}{OOM} \\
SynGen & cont. & 656.0 & 0.916 & $0.994$ & $0.000^\ast$ \\
\method & cont. & 328.7 & 0.981 & $0.981$ & $1.000^\ast$ \\

\bottomrule

\end{tabular}
}
\end{table}

\paragraph{Citeseer neighborhood memorization analysis.}\label{app:citeseer-nm}
Citeseer has 3{,}327 nodes and 4{,}552 undirected edges in 438 connected components. The largest component covers 63.7\% of the nodes, 16.4\% of the nodes are in components of at most two nodes, and 40.0\% have degree one. \method{} generates intra-community edges under a per-node degree budget inside the node's own leaf community. Let $C(i)$ be the leaf community of node $i$ and let $d_i$ be its number of neighbors inside $C(i)$. The generator selects those $d_i$ neighbors from the $|C(i)| - 1$ other members of the community, so the node admits $\binom{|C(i)| - 1}{d_i}$ distinct intra-community neighborhoods. When $d_i = |C(i)| - 1$, this count is one, and the reference neighborhood is the only one the model can produce. A two-node component is the extreme case. Each node has one candidate, so the reference edge is the only edge \method{} can place, and both nodes score a Jaccard similarity of 1.

Because \method{}'s neighborhood memorization on Citeseer is far above its value on the other four datasets, we analyze it further alongside CELL and DC-SBM. For each node, we compute the probability that a generator reproduces the reference neighborhood by random chance. A generator that picks $d_i$ of the $m_i = |C(i)| - 1$ candidates at random does so with probability $1/\binom{m_i}{d_i}$. We call a node constrained when this exceeds $0.05$. We then shuffle and rewire the reference graph itself while keeping each node's degree unchanged and each edge in the same pair of leaf communities. Isolated nodes are excluded, and we average over 20 shuffles. Of the 3{,}279 evaluated nodes, 1{,}064 are constrained. These rewired graphs reach a memorized-node fraction of $0.286 \pm 0.003$, above \method{}'s $0.183$. On the constrained nodes, the memorized-node fraction is $0.858$ for the rewired graphs and $0.561$ for \method{}. Leaf communities of at most two nodes contribute 82.9\% of \method{}'s memorized neighborhoods.

Splitting the nodes this way separates the three models. Over the 2{,}215 unconstrained nodes, \method{} reproduces 4 neighborhoods, DC-SBM 7, and CELL 299. Restricted to the unconstrained nodes of the largest component, \method{} reproduces 4 neighborhoods, DC-SBM 5, and CELL 292.

DC-SBM has a lower memorized-node fraction, $0.085$, but the whole difference lies in the constrained nodes, where \method{} scores $0.561$ and DC-SBM $0.257$. On the unconstrained nodes, both reproduce fewer than $10$ neighborhoods.

\subsection{Downstream Evaluation} \label{app:downstream}

We test whether a generated graph can serve as a substitute for the reference in node classification. Three settings are measured. TRTR trains and tests on the reference, setting the expected ceiling for the downstream task. TSTR trains on the generated graph and tests on the reference graph, measuring whether a model trained on generated data transfers to the reference. TSTS trains and tests on the generated graph, which measures how hard the generated task is.

Only three baselines generate attributes (GenCAT, SynGen, and GraphMaker, in two variants), and only these are included. We do not let models inherit reference attributes through the node alignment, since that would give them real features and an unfair advantage.

We use a $60/20/20$ train/validation/test split, drawn with a fixed seed. Every model is evaluated on the same split, and all models use the same training seeds. SynGen generates additional nodes, so its masks are drawn class by class to match the size and class distribution of the reference split.

Utility is the TSTR divided by TRTR, averaged over four two-layer architectures, GCN~\citep{GCN}, GraphSAGE~\citep{SAGE}, GAT~\citep{GAT}, and SGC~\citep{SGC}, and five seeds. For the architecture rankings, we train 20 configurations under both TSTS and TRTR: GCN, GraphSAGE, GAT, GIN~\citep{GIN}, and SGC at one to four layers, each over three seeds. Spearman and Pearson correlations are computed between the TSTS and TRTR accuracies of the 20 configurations. All runs use a hidden dimension of $64$, Adam optimizer at a learning rate of $0.01$ with weight decay of $5\times10^ {-4}$, and at most $200$ epochs, with early stopping on validation accuracy and a patience of $50$ epochs.

\subsubsection{Absolute accuracies}

Table~\ref{tab:downstream-abs} lists absolute accuracies next to the majority-class rate of each graph. The utility ratio alone hides class imbalance: when the reference accuracy is close to the majority rate, a model that predicts the majority class still achieves a high ratio. Flickr is such a case. The reference reaches $0.50$ against a majority rate of $0.42$, and every generated graph returns exactly the majority rate under both TSTR and TSTS. The task does not separate the models on Flickr, so we report no further downstream results for it. 

SynGen falls below the majority rate under both TSTR and TSTS on all four evaluation datasets, which is consistent with the design discussed in Section~\ref{sec:downstream}. Its TSTS on Cora ML, Amazon Photo, and Amazon Computers is also below the uniform chance.

\begin{table}[t]
\centering
\caption{Absolute node classification accuracy, averaged over architectures and seeds. The majority-class rate is the accuracy of always predicting the most frequent class.}
\label{tab:downstream-abs}
\small
\begin{tabular}{ll|ccccc}
\toprule
Model & & Citeseer & Cora ML & A. Photo & A. Computers & Flickr \\
\midrule
Majority class & & 0.21 & 0.30 & 0.27 & 0.37 & 0.42 \\
Reference & TRTR & 0.69 & 0.86 & 0.94 & 0.90 & 0.50 \\
\midrule
GenCAT & TSTR & 0.52 & 0.59 & 0.44 & 0.38 & 0.42 \\
 & TSTS & 0.71 & 0.81 & 0.72 & 0.41 & 0.42\\
GraphMaker S & TSTR & 0.71 & 0.85 & 0.80 & \multirow{2}{*}{TO} & \multirow{2}{*}{OOM} \\
 & TSTS & 0.62 & 0.83 & 0.99 \\
GraphMaker A & TSTR & 0.73 & 0.84 & 0.92 & \multirow{2}{*}{TO} & \multirow{2}{*}{OOM} \\
 & TSTS & 0.82 & 0.94 & 0.95 \\
SynGen & TSTR & 0.17 & 0.22 & 0.21 & 0.22 & 0.42 \\
 & TSTS & 0.19 & 0.02 & 0.04 & 0.03 & 0.42 \\
\method{} & TSTR & 0.59 & 0.70 & 0.76 & 0.69 & 0.42 \\
 & TSTS & 0.56 & 0.73 & 0.83 & 0.77 & 0.42 \\
\bottomrule
\end{tabular}
\end{table}

\subsubsection{Homophily and the TSTS gap}
\label{app:homophily}
TSTS trains and tests on the same generated graph. A generator that adds same-class edges makes this easier for message-passing models, which raises TSTS without bringing the graph closer to the reference. On Citeseer, Cora ML, and Amazon Photo, the three datasets on which GraphMaker completes, we measure how much of each TSTS$-$TRTR gap homophily explains. We use adjusted homophily~\citep{platonov2023},
\begin{equation*}
  h_{\mathrm{adj}} = \frac{h_{\mathrm{edge}} - \sum_c p_c^2}{1 - \sum_c p_c^2},
\end{equation*}
where $h_{\mathrm{edge}}$ is the fraction of edges that join nodes of the same class and $p_c = D_c / 2|E|$ is the degree share of class $c$, with $D_c$ the total degree of its nodes. 

For each target level of $h_{\mathrm{adj}}$, we rewire the reference with degree-preserving double edge swaps. The swap is valid only when it drives the number of same-class edges toward that level. With degrees and labels fixed, $h_{\mathrm{adj}}$ is a linear function of this number, so a target homophily fixes a target count. The targets are $0$, $h_{\mathrm{ref}}/2$, five steps from $h_{\mathrm{ref}}$ to $0.95$, and the $h_{\mathrm{adj}}$ of each generated model, over two rewiring seeds. Training the downstream models of Section~\ref{sec:downstream} on each rewired graph yields a curve $C(h)$ from homophily to accuracy. For a model with homophily $h_m$, $C(h_m) - \mathrm{TRTR}$ is the gain the reference itself shows at that homophily, and $\mathrm{TSTS} - C(h_m)$ is the excess that is not caused by homophily. We also rewire each generated graph to $h_{\mathrm{ref}}$, measure TSTS again, and train a logistic regression on the attributes alone.

GenCAT is excluded from this analysis. Part of its GraphSAGE and GAT fits collapse to the majority rate on all three datasets, so its TSTS is not a stable estimate.

Table~\ref{tab:homophily_properties} lists the homophily, edge count, and feature-only (without graph structure) accuracy of each graph, and Table~\ref{tab:homophily_calibration} splits each TSTS$-$TRTR gap into its homophily and the parts that cannot be explained without homophily. Whenever a generated graph is easier in the downstream task, its homophily is also higher than the reference. For these graphs, once the reference is rewired to the generated graph's homophily, the remaining difference is between $-0.033$ and $+0.005$. Which GraphMaker variant inflates homophily depends on the dataset: the asynchronous one on Citeseer and Cora ML, and both on Amazon Photo. On Citeseer and Cora ML, the inflating variant also generates two to four times as many edges as the reference. In contrast, the asynchronous variant on Amazon Photo raises homophily at $0.92$ of the reference edge count and still widens the gap, so the inflation does not depend on the extra edges alone.

GraphMaker also generates attributes that separate the classes better than the real ones. In all six cases, its feature-only accuracy is $0.05$ to $0.12$ above the reference, reaching $0.995$ for the synchronous variant on Amazon Photo. This makes the task easier independently of homophily. Rewiring the asynchronous variant to $h_{\mathrm{ref}}$ lowers its TSTS by only $0.01$ to $0.02$, and it stays $0.10$ above TRTR on Citeseer. The synchronous variant on Citeseer shows the attribute difference alone: generated homophily is below the reference, yet in the calibration run its TSTS matches TRTR (Table~\ref{tab:homophily_calibration}), with an excess of $+0.062$.

For \method{}, generated homophily is $0.16$ to $0.25$ below the reference and feature-only accuracy $0.16$ to $0.32$ below, and its TSTS is $0.07$ to $0.13$ below TRTR. Rewired to $h_{\mathrm{ref}}$, it comes within $0.05$ of TRTR. Its unexplained remainder ranges from $-0.001$ on Citeseer to $-0.084$ on Amazon Photo.

This analysis has three limitations. First, rewiring alone raises TSTS. A control that applies the same number of swaps at fixed homophily scores $0.013$ to $0.025$ above TRTR, so rewired differences smaller than that are within the effect of rewiring itself. Second, equal homophily does not mean equal structure. The targeted swaps distribute the same-class edges more evenly than in a real graph. Third, rewiring the reference keeps the real attributes, and rewiring a generated graph keeps the generated ones. The two results are therefore reflect different attributes and cannot be combined into one number.

\begin{table}[t]
\centering
\small
\setlength{\tabcolsep}{4pt}
\caption{Properties of the reference and generated graphs. $h_{\mathrm{adj}}$: adjusted homophily. Feature-only: accuracy of a logistic regression trained on node attributes without edges.}
\label{tab:homophily_properties}
\begin{tabular}{llccc}
\toprule
Dataset & Model & $h_{\mathrm{adj}}$ & $|E|/|E_{\mathrm{ref}}|$ & Feature-only \\
\midrule
\multirow{4}{*}{Citeseer}
& Reference    & 0.671 & 1.00 & 0.686 \\
& \method{}    & 0.512 & 1.00 & 0.529 \\
& GraphMaker S & 0.516 & 0.86 & 0.766 \\
& GraphMaker A & 0.870 & 4.28 & 0.810 \\
\midrule
\multirow{4}{*}{Cora ML}
& Reference    & 0.749 & 1.00 & 0.798 \\
& \method{}    & 0.503 & 1.03 & 0.482 \\
& GraphMaker S & 0.639 & 0.64 & 0.849 \\
& GraphMaker A & 0.926 & 2.17 & 0.844 \\
\midrule
\multirow{4}{*}{A. Photo}
& Reference    & 0.785 & 1.00 & 0.913 \\
& \method{}    & 0.586 & 0.97 & 0.621 \\
& GraphMaker S & 0.935 & 5.18 & 0.995 \\
& GraphMaker A & 0.900 & 0.92 & 0.968 \\
\bottomrule
\end{tabular}
\end{table}

\begin{table}[t]
\centering
\small
\setlength{\tabcolsep}{4pt}
\caption{Homophily calibration of TSTS. TSTS$-$TRTR is split into $C_{\mathrm{ref}}(h_m)-\mathrm{TRTR}$, the change in accuracy when the reference is rewired to the model's homophily $h_m$ (Table~\ref{tab:homophily_properties}), and the residual $\mathrm{TSTS}-C_{\mathrm{ref}}(h_m)$. $C_{\mathrm{gen}}(h_{\mathrm{ref}})$: TSTS after rewiring the generated graph to the reference homophily $h_{\mathrm{ref}}$. TRTR is $0.744$ on Citeseer, $0.857$ on Cora ML, and $0.938$ on Amazon Photo, each under a separate draw of the default split; on Citeseer, it differs from Table~\ref{tab:downstream-abs}, and so do the TSTS$-$TRTR values. Rewiring alone raises accuracy by $0.013$ to $0.025$, so rewired differences smaller than that are not interpretable.}
\label{tab:homophily_calibration}
\begin{tabular}{llcccc}
\toprule
& & & \multicolumn{2}{c}{Ref.\ rewired to $h_m$} & Gen.\ rewired \\
\cmidrule(lr){4-5}
Dataset & Model & TSTS$-$TRTR & $C_{\mathrm{ref}}(h_m)-$TRTR & TSTS$-C_{\mathrm{ref}}(h_m)$ & $C_{\mathrm{gen}}(h_{\mathrm{ref}})-$TRTR \\
\midrule
\multirow{3}{*}{Citeseer}
& \method{}    & $-$0.069 & $-$0.068 & $-$0.001 & $-$0.019 \\
& GraphMaker S & $-$0.005 & $-$0.067 & +0.062   & +0.070 \\
& GraphMaker A & +0.126   & +0.137   & $-$0.011 & +0.102 \\
\midrule
\multirow{3}{*}{Cora ML}
& \method{}    & $-$0.129 & $-$0.076 & $-$0.053 & +0.042 \\
& GraphMaker S & $-$0.026 & $-$0.028 & +0.003   & +0.029 \\
& GraphMaker A & +0.084   & +0.117   & $-$0.033 & +0.063 \\
\midrule
\multirow{3}{*}{A. Photo}
& \method{}    & $-$0.108 & $-$0.025 & $-$0.084 & +0.007 \\
& GraphMaker S & +0.054   & +0.048   & +0.005   & +0.051 \\
& GraphMaker A & +0.020   & +0.043   & $-$0.023 & +0.012 \\
\bottomrule
\end{tabular}
\end{table}

\subsection{Scalability}\label{app:scalability}

We measure scalability on the six scalability graphs of Table~\ref{tab:datasets}. Results are in Table~\ref{tab:app-scalability}. Most baselines do not run at this scale, so we compare against models that generate attributes and complete on Flickr, leaving GenCAT and SynGen. The other models that run on Flickr, including the classical ones, generate structure only and solve a smaller problem, so we leave them out.

We report training (or fitting) time and generation time separately. For GenCAT, fit time is the time required to estimate its parameters from the reference. For SynGen, fit time covers fitting the structure generator, the feature generator, and the aligner. For \method{}, the training time includes training across all three stages, and we also report the clustering time separately. Generation time is the wall-clock time to produce one graph. All runs use the hardware described in Section~\ref{sec:setup}.

\begin{table}[t]
\centering
\begin{threeparttable}
\caption{Training and generation time on the scalability graphs, in seconds (s), minutes (m), and hours (h).}
\label{tab:app-scalability}
\small
\begin{tabular}{ll|r|r|r|r|r|r}
\toprule
Model & & Flickr & Reddit & Yelp & OGBN & DGraphFin & IGB  \\
\midrule
GenCAT & Fit & 0.72 s & 11.55 s & 2.4 s$^\S$ & 20.34 s & 2.23 s & 30.92 s$^*$\\ 
 & Generate & 36.78 s & 33 m & 24.44 m$^\S$ & OOM & 19.91 m & OOM \\
 
SynGen & Fit & 14.03 s & 12.47 s & 261.33 s$^\ddagger$ & 50.30 s & 603.21 s & 332.59 s\\
& Generate & 5.68 s &  24.92 s & 18.38 s$^\ddagger$ & 16.18 s & 4.73 s & 53.9 s\\
 
\method{} & Clustering & 8.2 s & 423.3 s & 261.9 s & 151.86 m & 214.0 s & 31.91 m$^\dagger$ \\
& Train & 145.3 s & 22.71 m & 92.46 m & 5.69 h & 72.42 m & 5.16 h$^\dagger$ \\
& Generate & 1.2 s & 15.3 s & 107.5 s & 422.7 s & 23.1 s & 286.5 s$^\dagger$ \\
\bottomrule
\end{tabular}
\begin{tablenotes}[flushleft]\footnotesize
\item GenCAT runs on CPU only, and SynGen and \method{} use a single GPU. OOM for GenCAT means the run exhausted the 32~GB of host memory.
\item[$^\S$] GenCAT supports only single-label graphs, so on Yelp, we give it the first label of each node. Therefore, its Yelp input differs from that of the other models.
\item[$^*$] IGB Medium has no node labels, and GenCAT requires them, so we assign labels uniformly at random. GenCAT does not complete generation here, and its fit time on random labels may not be representative.
\item[$^\ddagger$] Yelp's 100 binary labels were written as 100 categorical columns for SynGen.
\item[$^\dagger$] \method{} runs on the largest connected component of IGB Medium ($98.8\%$ of its nodes), with leaf communities capped at $100{,}000$ nodes.
\end{tablenotes}
\end{threeparttable}
\end{table}

GenCAT fits in under a minute everywhere, but each sample then takes much longer to generate: $33$ minutes on Reddit and $20$ minutes on DGraphFin, and it exhausts host memory on OGBN Products and IGB Medium. \method{} spends most of its time before generation. Clustering and training together take up to $8.2$ hours, on OGBN Products, and generation takes between one and seven minutes on Yelp, OGBN Products, and IGB Medium and under 30 seconds on the others. One generated graph costs between $0.5\%$ and $2\%$ of the clustering and training time on every graph, so additional samples are cheap.

SynGen has the lowest total cost on every graph. It fits a small set of structural parameters and fits the attributes and their alignment separately, so its cost is closer to that of a classical generator than a learned one. Its fidelity and downstream results in Sections~\ref{sec:structural} and~\ref{sec:downstream} reflect the same design.

\subsection{Generating Graphs of Any Size} \label{app:anysize}

\method{} matches the reference node count by design, but its learned weights do not depend on $N$. The node count comes from the tree: each stage reads the leaves, their soft assignments, and the pooled matrices from disk in batches. Changing these inputs changes $N$ without retraining. 

We change the graph size by replicating or subsampling communities. For a scale factor $s$, the target node count is $sN$. If $s$ is an integer, the generated graph contains $s$ copies of every community. If $s > 1$ is not an integer, it contains $\lfloor s \rfloor$ copies of every community and a greedily chosen subset of additional communities. If $s < 1$, it contains only a greedily chosen subset.

\textbf{Stages 1 and 2.} Each leaf of each copy runs through Stage 1 and Stage 2 with its own reference inputs. Copies draw independent noise. Neither stage needs a change.

\textbf{Stage 3.} If Stage 3 were left unchanged, it would connect communities only within each copy, and the result would be a disjoint union of copies, disconnected even when the reference is connected. We therefore generate Stage 3 edges in two steps. First, Stage~3 runs inside each copy exactly as in the standard generation. Second, each of these edges is reassigned to a random pair of copies of its two endpoints, drawn from two independent random permutations of the copies. This keeps every node's degree, and changes only which copies of an edge join.

Two reassigned edges can land on the same pair of copies, and the duplicate is dropped, which lowers the edge count. The loss depends on the scale factor and is largest at scale factors 2 and 2.5 in Table~\ref{tab:anysize}.

We run this study on Flickr, sweeping scale factors of 0.5, 1, 1.5, 2, 2.5, 5, 10, and 100. Table~\ref{tab:anysize} reports the expected and generated node counts and two edge ratios: against the reference, and against the graph \method{} generates at scale one. Each entry is the mean and standard deviation over 3 samples.

\begin{table}[t]
\centering
\caption{Generation at different scale factors on Flickr. Scale is the scale factor $s$, $N_{\text{exp}}$ is the expected number of nodes, and $N_{\text{gen}}$ is the generated number of nodes. Edge rat.\ (gen) is the number of generated edges divided by $s$ times the edge count \method{} generates at scale one. Edge rat.\ (ref) divides it by $s$ times the reference edge count.}
\label{tab:anysize}
\begin{tabular}{rrrrr}
\toprule
Scale & $N_{\text{exp}}$ & $N_{\text{gen}}$ & Edge rat. (gen) & Edge rat. (ref)  \\
\midrule
0.5 & 44{,}625 & 45{,}225 & $1.040 \pm 0.044$ & $0.864 \pm 0.036$\\
1 & 89{,}250 & 89{,}250 & $1.000 \pm 0.001$ & $0.831 \pm 0.001$\\
1.5 & 133{,}875 & 135{,}058 & $0.992 \pm 0.014$ & $0.824 \pm 0.012$\\
2 & 178{,}500 & 178{,}500 & $0.987 \pm 0.001$ & $0.820 \pm 0.001$\\
2.5 & 223{,}125 & 223{,}433 & $0.985 \pm 0.005$ & $0.818 \pm 0.004$\\
5 & 446{,}250 & 446{,}250 & $0.992 \pm 0.000$ & $0.824 \pm 0.000$ \\
10 & 892{,}500 & 892{,}500 & $0.995 \pm 0.000$ & $0.827 \pm 0.000$ \\
100 & 8{,}925{,}000 & 8{,}925{,}000 & $0.999 \pm 0.000$ & $0.830 \pm 0.000$ \\
\bottomrule
\end{tabular}
\end{table}

For integer scale factors, the generated node count is exact. For fractional ones, it deviates slightly because communities are added whole. Edge rat.\ (gen) stays within 1.5\% of one for every scale factor above one and approaches one as the scale grows, since the fraction of colliding edges falls. At $s = 0.5$, it is $1.040$, because the chosen subset holds more nodes than the target. Edge rat.\ (ref) stays near $0.83$, which is the ratio \method{} already produces at scale one. Hence, the scale factor does not change the edge ratio relative to the reference. At scale 100, generation completes, producing a graph with 8.9M nodes and 37.3M edges.

\subsection{Ablation Study}
\label{app:ablation}

We run ablation studies on Cora ML because the sweeps produce many generated graphs, and Cora ML is one of the smallest references. We do not use Citeseer: its many small components fall into separate leaves whose candidate sets are too small for the edge-stage variants to differ. Cora ML's leaves are large enough to exercise both edge stages. We do not repeat the studies on the larger graphs.

We report the $W_1$ degree distance, the triangle ratio, and the Inter/Intra ratio for structural fidelity, edge overlap, and self-overlap for memorization, and TSTS$-$TRTR for downstream utility.

Each ablation changes one part of the pipeline while keeping the rest at the same configuration.

\subsubsection{Stage-wise ablation} 
Each stage draws noise during generation, and each can be switched off independently, yielding eight combinations. With no noise, Stage 2 encodes the observed adjacency of each leaf, and Stage 3 reads the observed inter-community edges, so the result follows the reference. Noise in all three stages is the configuration reported throughout the paper.

\subsubsection{Partitioning}
We replace Leiden with Louvain~\citep{Louvain} in both the leaf partition and the recursive tree construction, and separately with a random partition. Random partitioning keeps the community sizes Leiden produced and assigns nodes to them at random, so the leaves have the same sizes but no structural meaning.

\subsubsection{Hard assignment}
This ablation replaces the soft assignments with one-hot matrices over the same leaf partition. No node then carries mass in more than one child, and no bridge nodes will be found. Making the candidate pool drawn at random. The pooled connectivity and the cluster-affinity term remain. 

\subsubsection{Branching cap}
The branching cap $b$ bounds how many supernodes a tree node holds before the recursion splits it again. At $b = 2$, the recursion builds a binary tree. At $b = 1000$, it stops at the root, since Cora ML has fewer leaves than that, so every leaf community is a direct child of the root, and the hierarchy has one level.

\subsubsection{Joint attribute generation}
At generation, we replace the attributes produced by Stage 1 with the reference attributes, to test whether the edge stages depend on the generated attributes.

\subsubsection{Stage architectures}
We replace the architecture of one stage at a time. Stage 1 uses an MLP in place of the Transformer, so the attributes of nodes in the same leaf are generated independently. Stage 2 uses an inner-product decoder or an MLP edge predictor. Stage 3 uses an attention scorer in place of the bilinear scorer. The bilinear scorer is already the simpler option, so this row tests whether a more complex scorer helps. 

\subsubsection{Cluster affinity}
Stage 3 adds a cluster-affinity term to each pair score from the hierarchy. We compare three settings. Without the term, the scorer learns from the observed inter-community edges alone. In the reported setting, the term is added to the score with a learned weight, so the model starts from the connectivity that hierarchy contains. In the third setting, the term is a KL target during training and is not added to the scores at generation.

\begin{table}[b]
\centering
\small
\setlength{\tabcolsep}{4pt}
\caption{Stage-wise ablation on Cora ML. A check mark means the stage draws noise at generation. The last row is the configuration reported in the main text. SO is undefined in the first row, since all samples are identical without noise.}
\label{tab:abl-stagewise}
\begin{tabular}{ccc|rrr|rr|r}
\toprule
\multicolumn{3}{c|}{Noise} & \multicolumn{3}{c|}{Structure} & \multicolumn{2}{c|}{Memorization} & Downstream \\
\cmidrule(lr){1-3}\cmidrule(lr){4-6}\cmidrule(lr){7-8}\cmidrule(lr){9-9}
Node & Intra & Inter & Deg.\ W1 & Tri.\ Rat. & Inter/Intra & EO & SO & TSTS$-$TRTR \\
\midrule
& & & $1.502$ & $1.171$ & $0.871$ & $0.149$ & \text{n/a} & $-0.080\pm 0.012$\\
\checkmark & & & $0.590 \pm 0.062$ & $0.417 \pm 0.051$ & $0.837 \pm 0.035$ & $0.113$ & $0.104$ & $-0.118 \pm 0.019 $\\
& \checkmark & & $1.170 \pm 0.013$ & $0.095 \pm 0.004$ & $0.992 \pm 0.060$ & $0.112$ & $0.105$ & $-0.082 \pm 0.013 $\\
\checkmark & \checkmark & & $1.177 \pm 0.018$ & $0.091 \pm 0.005$ & $0.986 \pm 0.064$ & $0.111$ & $0.105$ & $-0.126\pm 0.024$ \\
& & \checkmark & $1.587 \pm 0.045$ & $1.360 \pm 0.048$ & $0.989 \pm 0.028$ & $0.074$ & $0.183$ & $-0.095\pm 0.013$ \\
\checkmark & & \checkmark & $1.011 \pm 0.011$ & $0.577 \pm 0.048$ & $0.899 \pm 0.087$ & $0.044$ & $0.066$ &  $-0.126\pm 0.020$ \\
& \checkmark & \checkmark & $1.038 \pm 0.030$ & $0.246 \pm 0.009$ & $0.902 \pm 0.087$ & $0.037$ & $0.061$ & $-0.097\pm 0.013$ \\
\checkmark & \checkmark & \checkmark & $1.075 \pm 0.024$ & $0.275 \pm 0.019$ & $1.030 \pm 0.083$ & $0.038$ & $0.066$ & $-0.125 \pm 0.029$ \\
\bottomrule
\end{tabular}
\end{table}

\begin{table}[t]
\centering
\small
\setlength{\tabcolsep}{4pt}
\caption{Component ablations on Cora ML. Each row changes one part of the pipeline and leaves the rest at the reported configuration, given in the first row.}
\label{tab:abl-components}
\resizebox{\columnwidth}{!}{
\begin{tabular}{ll|rrr|rr|r}
\toprule
& & \multicolumn{3}{c|}{Structure} & \multicolumn{2}{c|}{Memorization} & Downstream \\
\cmidrule(lr){3-5}\cmidrule(lr){6-7}\cmidrule(lr){8-8}
Ablation & Setting & Deg.\ W1 & Tri.\ Rat. & Inter/Intra & EO & SO & TSTS$-$TRTR \\
\midrule
\multicolumn{2}{l|}{\textbf{Reported}} & $1.075 \pm 0.024$ & $0.275 \pm 0.019$ & $1.030 \pm 0.083$ & $0.038$ & $0.066$ & $-0.125 \pm 0.029$ \\
\midrule
\multirow{2}{*}{Partitioning}
 & Louvain & $0.933 \pm 0.037$ & $0.277 \pm 0.015$ & $1.263 \pm 0.088$ & $0.037$ & $0.061$ & $-0.107 \pm 0.021$ \\
 & Random & $4.026 \pm 0.034$ & $3.515 \pm 0.069$ & $7.516 \pm 0.472$ & $0.017$ & $0.170$ & \text{n/a}\\
\midrule
Assignment
 & Hard & $1.226 \pm 0.026$ & $0.208 \pm 0.007$ & $0.838 \pm 0.048$ & $0.033$ & $0.044$ & $-0.142\pm 0.027$\\
\midrule
\multirow{2}{*}{Branching cap}
 & $b = 2$ &  $1.365 \pm 0.027$ & $0.332 \pm 0.015$ & $0.946 \pm 0.032$ & $0.036$ & $0.071$ & $-0.127\pm 0.026$\\
 & $b = 1000$ & $0.842 \pm 0.019$ & $0.278 \pm 0.007$ & $1.044 \pm 0.029$ & $0.039$ & $0.069$ & $-0.124\pm 0.027$\\
\midrule
Attribute source
 & Reference & $1.036 \pm 0.020$ & $0.238 \pm 0.011$ & $0.942 \pm 0.148$ & $0.037$ & $0.062$ & $-0.134\pm 0.024$\\
\midrule
\multirow{4}{*}{Architecture}
 & Stage 1, MLP & $1.128 \pm 0.033$ & $0.284 \pm 0.013$ & $0.864 \pm 0.099$ & $0.038$ & $0.067$ & $-0.135\pm 0.028$ \\
 & Stage 2, inner product & $1.252 \pm 0.050$ & $1.042 \pm 0.042$ & $1.073 \pm 0.118$ & $0.071$ & $0.127$ & $-0.129\pm 0.025$ \\
 & Stage 2, MLP & $1.703 \pm 0.051$ & $1.196 \pm 0.061$ & $0.930 \pm 0.054$ & $0.079$ & $0.131$ & $-0.121\pm 0.024$\\
 & Stage 3, att. & $0.917 \pm 0.032$ & $0.181 \pm 0.007$ & $0.827 \pm 0.137$ & $0.037$ & $0.051$ & $-0.134\pm 0.027$\\
 \midrule
\multirow{2}{*}{Affinity}
 & KL & $1.305 \pm 0.037$ & $0.238 \pm 0.011$ & $1.083 \pm 0.042$ & $0.033$ & $0.048$ & $-0.135\pm 0.025$\\
 & None & $1.432 \pm 0.021$ & $0.353 \pm 0.016$ & $1.247 \pm 0.059$ & $0.033$ & $0.056$ & $-0.132\pm 0.025$\\
\bottomrule
\end{tabular}
}
\end{table}

\subsubsection{Results}
Table~\ref{tab:abl-stagewise} separates the three stages. Edge overlap falls from $0.149$ with no noise to $0.038$ with noise in all stages, and the drop comes from Stage 3: the four combinations that read the observed inter-community edges sit between $0.111$ and $0.149$, and the four that generate them sit between $0.037$ and $0.074$. Triangle counts move in the opposite direction and with the intra-edge stage. When that stage encodes the observed adjacency of each leaf, the triangle ratio is between $0.417$ and $1.360$. When it draws from the prior instead, it is between $0.091$ and $0.275$. The model reproduces the triangle density of the reference only when it reads the reference adjacency. This is the trade-off from Section~\ref{sec:structural}, measured within our own pipeline instead of across models.

Node noise reaches the structure only through the encoder. With Stage 2 in its deterministic mode, switching node noise on changes the degree distance from $1.502$ to $0.590$ and the triangle ratio from $1.171$ to $0.417$. With Stage 2 generating and Stage 3 deterministic, the same switch changes both by less than $0.01$. The encoder reads the attributes, and the prior does not, so once Stage 2 samples from the prior, Stage 1 has little effect on the structure.

Table~\ref{tab:abl-components} shows an ablation study on components. Replacing Leiden with random communities of the same sizes changes the degree distance from $1.075$ to $4.026$, the triangle ratio from $3.515$ to $3.515$, and the Inter/Intra ratio from $7.516$ to $7.516$. Louvain, which optimizes the same objective, gives a lower degree distance ($0.933$) but a higher Inter/Intra ratio ($1.263$). The cluster-affinity term also matters: without it, the Inter/Intra ratio rises from $1.030$ to $1.247$ and the degree distance to $1.432$. Using it as a KL target instead of adding it to the scores recovers part of this.

The remaining components have smaller effects on this graph. Hard assignments reach an Inter/Intra ratio of $0.838$ against $1.030$ with soft ones, so removing bridge nodes that Stage 3 draws its endpoints from degrades the result without breaking it for this dataset. A binary tree and a single-level hierarchy both stay within $0.09$ of the reported Inter/Intra ratio. Hence, the depth of the recursion matters little on a graph of this size, although the binary tree raises the degree distance to $1.365$. Replacing the Transformer of Stage 1 with an MLP keeps the degree distance, the triangle ratio, and the downstream gap close to the reported values and lowers the Inter/Intra ratio to $0.864$, within about two standard deviations. Replacing the bilinear scorer of Stage 3 with an attention scorer lowers the degree distance to $0.917$ but moves the Inter/Intra ratio further from one, to $0.827$, and lowers the triangle ratio to $0.181$.

The two simpler decoders for Stage 2 show the same trade-off as the stage-wise table. An inner-product decoder reaches a triangle ratio of $1.042$ and an MLP predictor $1.196$, both closer to the reference than the $0.275$ of the reported configuration, but their edge overlap rises to $0.071$ and $0.079$ and their self-overlap to $0.127$ and $0.131$. They recover the triangle density by staying closer to the reference edge set.

Replacing the generated attributes with the reference attributes changes the structural metrics by at most about two standard deviations. At generation, Stage 2 draws its latents from the prior, and Stage 3 reads memberships and pooled connectivity, so neither edge stage reads the generated attributes. Attributes and edges are tied through the shared soft memberships, not through the realized attributes.

\end{document}